\documentclass[10pt,twocolumn,letterpaper]{article}

\usepackage[pagenumbers]{cvpr} % To force page numbers, e.g. for an arXiv version
\usepackage{booktabs}        % For \toprule, \midrule, \bottomrule
\usepackage{multirow}        % For multirow cells
\usepackage{graphicx}        % For \resizebox and \rotatebox
\usepackage[table]{xcolor}          % For \cellcolor and colored rows
\usepackage{colortbl}        % For colored table cells
\usepackage{amssymb}         % For \checkmark and \xmark
\usepackage{amsmath}         % For math symbols like \mathcal
\usepackage{makecell}        % For \shortstack inside table cells
\usepackage{adjustbox}       % Optional, for better control of resizing tables
\usepackage{rotating}        % For rotated text (\rotatebox alternative)

\definecolor{tabhighlight}{rgb}{0.88,0.95,1}
\definecolor{negativehighlight}{rgb}{1,0.95,0.91}
\definecolor{trainedintra}{RGB}{230, 240, 255}    % Light blue
\definecolor{trainedinter}{RGB}{230, 255, 240}   % Light green
\definecolor{tabhighlightbluetext}{rgb}{0.2,0.4,0.8}

\definecolor{tabhighlightpurple}{rgb}{0.95,0.88,1}
\definecolor{tabhighlightpurpletext}{rgb}{0.7,0.4,0.8}

\definecolor{ForestGreen}{RGB}{34, 139, 34}

\usepackage{float}
\usepackage{dsfont}

\definecolor{cvprblue}{rgb}{0.21,0.49,0.74}
\usepackage[pagebackref,breaklinks,colorlinks,allcolors=cvprblue]{hyperref}

\def\paperID{-----} % *** Enter the Paper ID here
\def\confName{CVPR}
\def\confYear{2026}

\title{Reevaluating the Intra-Modal Misalignment Hypothesis in CLIP}
\author{Jonas Herzog\\
Zhejiang University\\
{\tt\small jherzog@zju.edu.cn}
\and
Yue Wang\\
Zhejiang University\\
{\tt\small wangyue@iipc.zju.edu.cn}
}

\begin{document}
\maketitle
\begin{abstract}
Recent research suggested that the embeddings produced by CLIP-like contrastive language-image training are suboptimal for image-only tasks.
The main theory is that the inter-modal (language-image) alignment loss ignores intra-modal (image-image) alignment, leading to poorly calibrated distances between images.
In this study, we question this intra-modal misalignment hypothesis.
We reexamine its foundational theoretical argument, the indicators used to support it, and the performance metrics affected.
For the theoretical argument, we demonstrate that there are no such supposed degrees of freedom for image embedding distances.
For the empirical measures, our findings reveal they yield similar results for language-image trained models (CLIP, SigLIP) and image-image trained models (DINO, SigLIP2).
This indicates the observed phenomena do not stem from a misalignment specific to the former.
Experiments on the commonly studied intra-modal tasks retrieval and few-shot classification confirm that addressing task ambiguity, not supposed misalignment, is key for best results.
{\small Project page: \url{https://vision-kek.github.io/Is-CLIP-Really-Misaligned/}}.
\end{abstract}    
\section{Introduction}
\label{sec:intro}
%gpt
The success of contrastive language-image pretraining exemplified by CLIP \cite{clip} has significantly shaped the contemporary era of multimodal, foundational models.
The most fundamental and still ubiquitous use-case is to embed images and texts in a shared space, and then compare them via cosine similarity.
Following this simple approach, many long-standing computer vision problems from classification to detection \cite{owlvit,owlv2} to segmentation \cite{maskclip, ovseg} became addressable in the popular open-vocabulary zero-shot setting.

\begin{figure}[htbp]
    \centering
    \includegraphics[width=\linewidth]{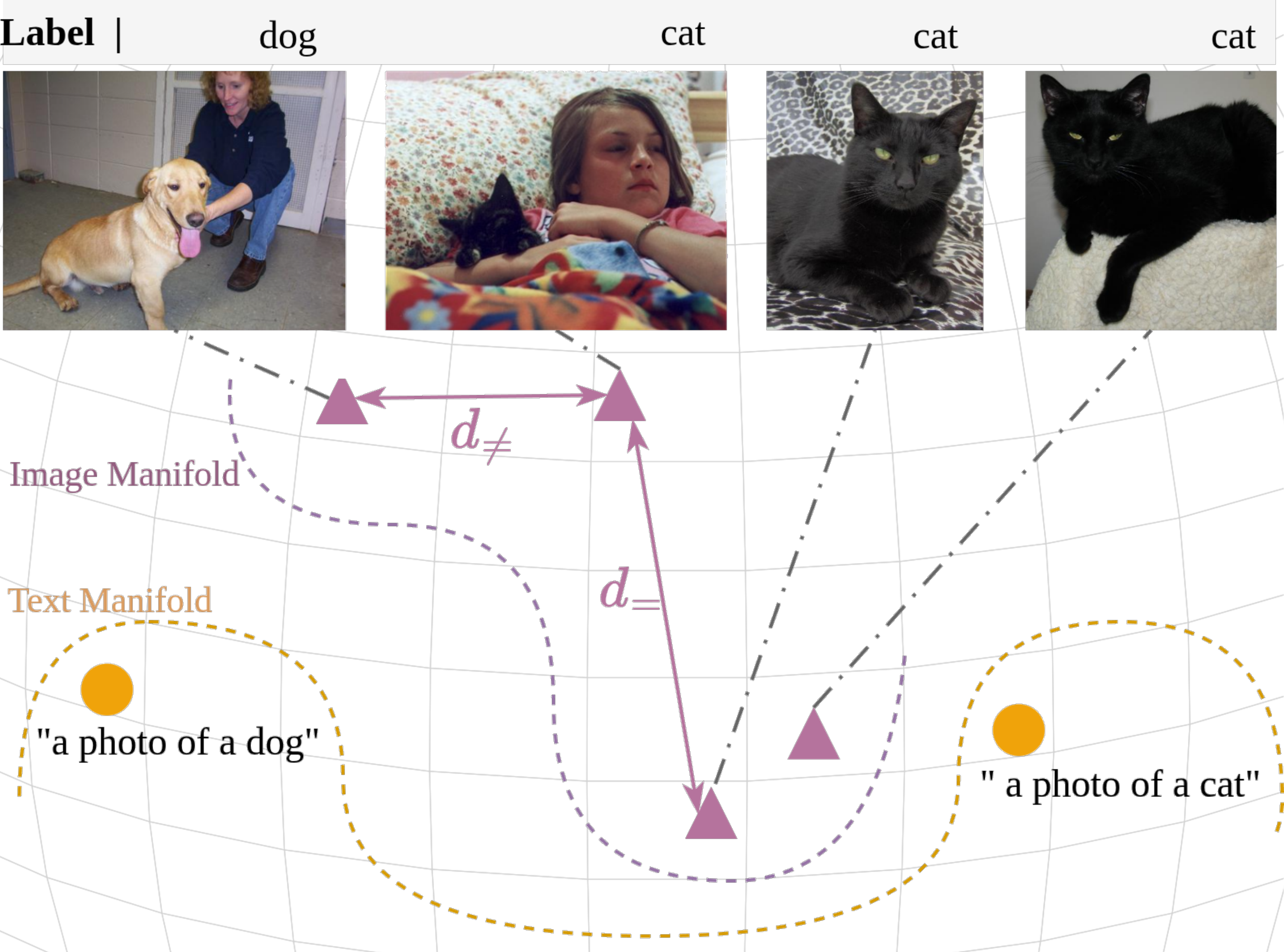}
    % The motivation of \textit{Cross the Gap} \cite{ctg}.
    \caption{Previous work illustrated an intra-modal misalignment in CLIP space by showing there are cat images closer to a dog ($d_{\neq}$) than to another cat ($d_{=}$). We argue $d_{\neq} < d_=$ is \textit{no} sign of misalignment. For a labeled downstream dataset, intra-class variance of open vocabulary models is \textit{expected and desired} to capture semantics and style beyond the narrow dataset-specific labels. Classifying and retrieving with frozen CLIP image embeddings still works well when similarities are measured along the dataset-specific semantic axes. Here the horizontal axis captures dog/cat.}
    \label{fig:introfig}
\end{figure}

Nevertheless, recent works have raised concerns that these embeddings may not be ideal for image-only tasks. 
The so-called \textit{intra-modal misalignment hypothesis} argues that contrastive language–image training focuses solely on inter-modal (image–text) alignment while neglecting intra-modal (image–image) alignment \cite{ctg, susx, Barbier2025BridgingTM, coder}.
As a result, even though the model produces good image-text similarities, the same might not be true for image-image similarities.
This effect is often illustrated as in \cref{fig:introfig}.

The hypothesis relies on the following pillars:
(i) a theoretical argument that misalignment emerges from unconstrained degrees of freedom in the embedding space, and
(ii) empirical evidence through proposed misaligment indicators such as cosine similarity histograms, modality-gap magnitudes, and retrieval or few-shot classification performance metrics.
\\
In this work, we critically re-examine these arguments.

For the theoretical model, we demonstrate that the degree of freedom argument breaks when considering a sufficiently large set of embeddings.
Image–image similarities can in fact be recovered entirely from image–text similarities.
This shows that image–image structure is not arbitrary but a consequence of the learned image-text structure.

For the empirical indicators that were proposed to reveal the misalignment, we experiment with these measures and find they are no reliable indicators of intra-modal alignment quality.
For example, we argue intra-class variance as shown in \cref{fig:introfig} is not a sign of misalignment.
Instead, this is normal behavior because without fine-tuning, open-vocabulary models are expected to capture semantics and styles beyond the closed-vocabulary of a narrow downstream dataset.
More crucially, the same indicators would also suggest ``misalignment'' in non-CLIP vision encoders such as DINO \cite{dino}, which have never been trained with language supervision — revealing that these signals are artifacts of the metrics rather than of the training objective.

Based on our findings, we propose a simple alternative method that measures similarities in class-relevant axes.
We conduct evaluations on retrieval and few-shot classification, which have been previously assumed to be affected by the misalignment, and provide a reinterpretation of the results.
Finally we discuss the role of the modality gap \cite{Liang2022MindTG} which is often cited \cite{ctg, coder, RetrievalAugmentedTaskAdaptionForVLMs} in the context of intra-modal misalignment.
In summary:
\begin{itemize}
    \item We critically re-examine the intra-modal misalignment hypothesis, which directly impacts all methods that use CLIP or SigLIP to compare image embeddings with each other.
    \item We point out the limitations of previous arguments and evidences for such misalignment and provide an alternative explanation for their findings.
    \item Consequently, we construct a simple alternative method that confirms that the best performance on the previously studied few-shot classification and retrieval tasks can be achieved without assuming a misalignment in CLIP.
\end{itemize}
\section{The Intra-Modal Misalignment Hypothesis}
\label{sec:hypothesis}
%\subsection{Definition}
%Following  and \cite{ctg}, the intra-modal misalignment hypothesis states:
%Two core excerpts capture the core of the intra-modal misalignment hypothesis:
%The intra-modal misaligment hypothesis posits:
We examine the following belief:
\begin{quote}
    %An inter-modal contrastive loss as in CLIP does not enforce any intra-modal constraints \cite{ctg}. %Intra-modal similarities are disregarded \cite{susx}.
    CLIP is not optimized for uni-modal scenarios \cite{coder}.
    This leads to intra-modal misalignment \cite{ctg} / miscalibration \cite{susx}, its performance in intra-modal tasks is not guaranteed \cite{coder}.
    %Resulting image-image similarities may not reflect the true image similarities \cite{susx}. 
    Only exploiting the image encoder of such models is highly suboptimal for intra-modal tasks like image-to-image retrieval \cite{ctg}.
\end{quote}

\subsection{The Supporting Evidence}
\label{subsec:supporting}
\begin{figure}[ht]
    \centering
    \includegraphics[width=\linewidth]{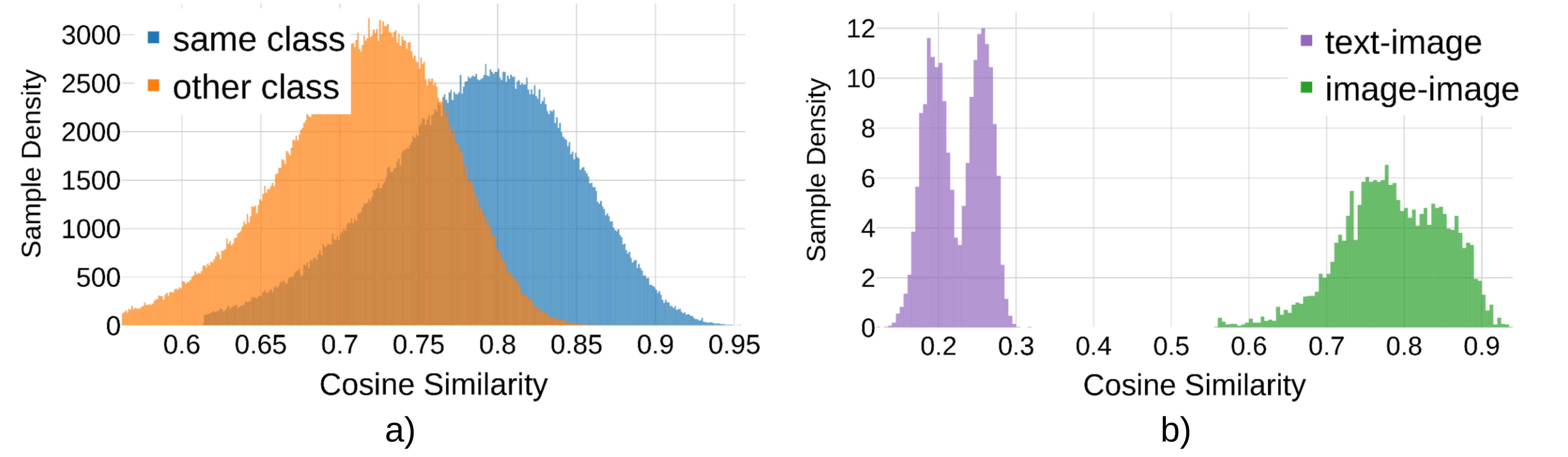}
    \caption{Pairwise cosine similarity distributions.
    \textit{Left}: Similarities between same class (blue) and opposite class (orange) image feature pairs. A high overlap ratio between the two colors was previously highlighted as an indicator for an intra-modal misalignment issue in CLIP.
    \textit{Right}: Similarity distributions of image-text pairs (purple) versus image-image pairs (green). Because CLIP is only supervised on the former, the divergence has previously prompted concerns about whether the latter reflect true similarities. CLIP ViT-B/16. Dataset as in \cref{tab:catdog_retrieval}.
   }
    \label{fig:cossimhist_prev}
\end{figure}

We review some evidence in favor of the above intra-modal misalignment hypothesis.

The consequence of having only a cross-modal loss term in CLIP has been \textbf{theoretically} explained \cite{susx, coder} by degrees of freedom in the embedding space.
This idea has been illustrated similar to \cref{fig:dof}a-c.

Further, to measure and expose poorly calibrated similarities, several \textbf{indicators} were proposed.
The main tool for this purpose are cosine similarity histograms, where the distribution of pairwise similarities between images are recorded.
There are two types of similarities that are examined, by class and by modality, shown in \cref{fig:cossimhist_prev}.
%The first compares the distribution of cosine similarities between classes, the second between modalities.
For the similarity distributions by classes \cite{logitsdeconfusion,ctg}, the point is straightforward:
Two images of the same class should have higher similarity than two images of different classes.
As in \cref{fig:cossimhist_prev}a, if the two histograms that capture the similarities of same-class and different-class image pairs have a high overlap, it is hypothesized this indicates poor class separation and hence is a sign of misalignment.
On the other hand, measuring the similarity distributions by modality \cite{susx, Barbier2025BridgingTM,semobridge} as in \cref{fig:cossimhist_prev}b reveals that image embeddings have a much higher similarity to each other than to text embeddings.
This is considered problematic especially for few-shot classification in the vision-language model adaptation literature \cite{susx, Barbier2025BridgingTM, coder, semobridge} because these work combine inter-modal and intra-modal similarity scores, such that it is questioned if the two differently distributed scores can be treated equally.

This finding of higher intra-modal similarities is a consequence of the modality gap \cite{Liang2022MindTG}, which describes the effect that image embeddings cluster in a manifold clearly separated from the text embedding cluster.
Distances across the cluster are therefore naturally higher on average than within.
Some work \cite{ctg, Barbier2025BridgingTM} has also attributed the misalignment to the modality gap, but the existing experiments \cite{Liang2022MindTG, ctg} documented that a shifting or fine-tuning approach to narrow the gap tends to worsen performance.

\begin{figure}
    \centering
    \includegraphics[width=\linewidth]{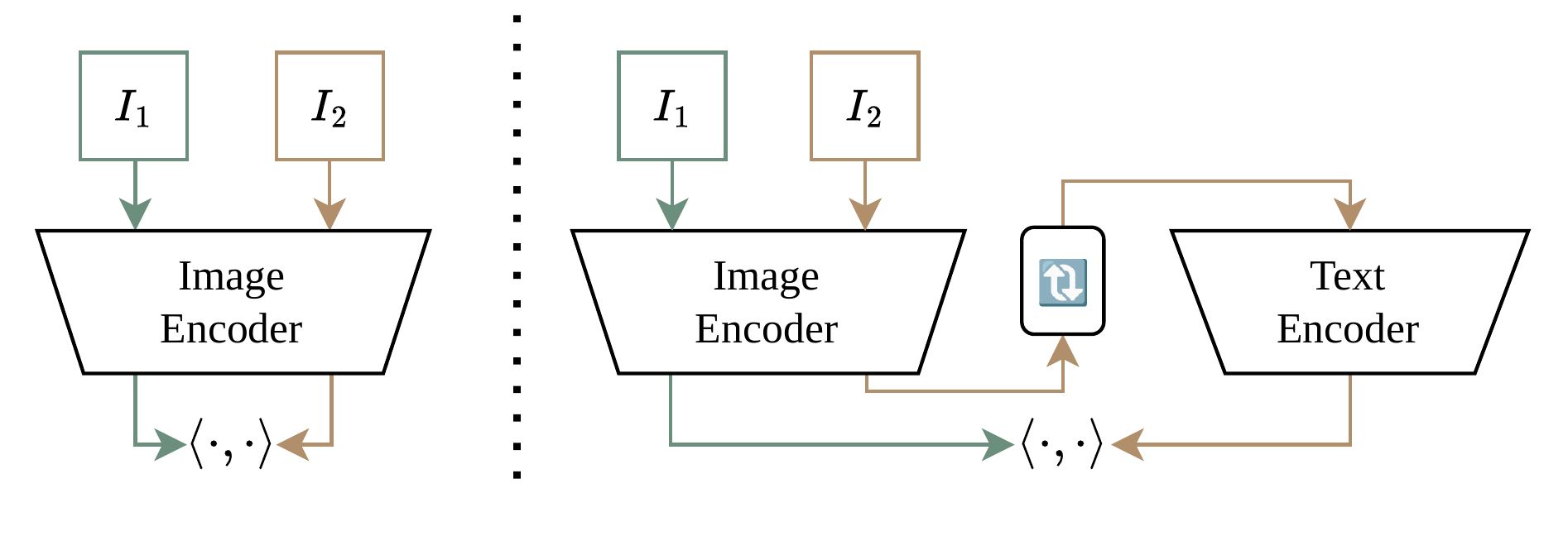}
    \caption{Motivated by the intra-modal misalignment hypothesis, previous work \cite{ctg} posited it is necessary to convert image-image comparison (left) into image-text comparison (right).} 
    \label{fig:oti}
\end{figure}

To mitigate the assumed misalignment, prior works then proposed methods for better \textbf{performance}.
The most common idea is to avoid measuring image-image similarities in favor of text-image similarities.
For few-shot classification, Tip-X \cite{susx} proposed to address uncalibrated image-image embedding distances in few-shot classification by using image-text similarities as a bridge.
Instead of directly comparing image features, it constructs ``signatures'' for both test and few-shot images by measuring their affinity with text classifier weights. The final affinity between images is then calculated as the KL-divergence between these inter-modal probability distributions.
With a similar motivation, Mistretta \etal \cite{ctg} proposed to tune a text token that represents the image information, an approach illustrated in \cref{fig:oti} and dubbed Optimization-based Textual Inversion (OTI) inspired by \cite{Gal2022AnImageIsWorthOneWord}.
This proved to be more efficient on image-image retrieval than comparing image embeddings directly.
In another work, Barbier \etal \cite{Barbier2025BridgingTM} found that simply normalizing the cosine similarity distributions as in \cref{fig:cossimhist_prev}a to equal mean and variance can bring improvements for few-shot object detection with CLIP-like VLMs.
Chao \etal \cite{coder} also assumed wrong distances between images for intra-modal tasks, and therefore introduced to represent an image by its cross-modal distances to text generated with the help of large language models.

In summary, a substantial body of prior work has interpreted calibrated similarity statistics, modality-wise clustering, and improved performance from text-based surrogates as evidence of intra-modal misalignment.

\subsection{The Prior Opposing Indicators that Call for Reassessment} % Sources of Doubt and Necessity to Reassess
\label{subsec:opposing}

While the preceding section reviewed evidence supporting the intra-modal misalignment hypothesis, there also exist several observations that lower the prior probability that the hypothesis is true before considering the evidence presented in our paper.

A first source of skepticism comes from some influential literature that has successfully employed CLIP and related vision–language models for uni-modal tasks such as image classification \cite{Geirhos2024TowardsFP}, few-shot adaptation \cite{tipadapter, Wang2024AHB, proker}, image generation \cite{Ruiz2022DreamBoothFT}, and video synthesis \cite{Esser2023StructureAC, Zhang2024VideoDiffusion}. Further, \cite{RetrievalAugmentedTaskAdaptionForVLMs} report that image-to-image retrieval actually outperforms image-to-text retrieval in downstream VLM adaptation tasks. If CLIP embeddings were indeed severely misaligned within the visual modality, it would be difficult to reconcile this with the strong results reported in these works, particularly those experiments that exclusively rely on the image encoder. 
A second reason to question the misalignment interpretation concerns the meaning of intra-class variance.
An alternative explanation for the phenomena illustrated in \cref{fig:introfig} has been articulated in \cite{ape}: embeddings naturally contain both task-specific and task-irrelevant components. From this perspective, high intra-class variance may reflect the coexistence of other semantic factors rather than represent a flaw in the embedding space. The challenge, therefore, is not to ``correct'' image–image distances, but to identify task-relevant axes (classifier) given the limited samples \cite{Lin2023MultimodalityHU, Farina2025RethinkingFA}.

Taken together, \cref{subsec:supporting} and \cref{subsec:opposing} highlight a tension: while many works argue for intra-modal misalignment, several existing results in the broader literature are difficult to reconcile with the misalignment view.
This inconsistency calls for a careful reassessment of the hypothesis.

\subsection{A Preliminary Experiment}
As a first step toward reassessing the misalignment hypothesis, we revisit a simple experiment in \cite{ctg} which particularly inspired this work.
In this experiment, poor CLIP retrieval performance on a very easy cats and dogs legacy dataset \cite{dogsvscats} was attributed to the intra-modal misalignment.

\begin{table}
\caption{Repeating the demonstrative experiment in \cite{ctg} on the simplistic Dogs vs Cats legacy dataset, where near-perfect results are expected.
It was suggested in \cite{ctg} that poor results with CLIP image-image similarities evidences a misalignment in the image-image space.
This hypothesis gets no evidence when swapping model for the uni-modal DINO, a widely acknowledged state-of-the-art image embedder.
CLIP scores highest, suggesting that the observed low metrics are not caused by a model weakness and hence neither by a misalignment.
Instead, the performance gap between text-image (T-I) and image-image (I-I) can be attributed to the ambiguity \cite{Lin2023MultimodalityHU} in the way the task is conveyed to the model: Two images with opposite labels might still share enough other concepts to be justifiably similar.}
%, in which visual concept in the comparative image(s) is decisive for the dataset-specific label. %in \cite{ctg}.
% An image contains multiple concepts, and is only one variation
\label{tab:catdog_retrieval}
\centering
 \resizebox{1.0\linewidth}{!}{
\begin{tabular}{ccccc}
    \toprule
    & \multicolumn{2}{c}{Retrieval} & \multicolumn{2}{c}{Classification} \\
    \midrule
    Model & $T\text{-}I$ & $I\text{-}I$ & $T\text{-}I$ (0-shot) & {$I\text{-}I$ (1 \textbar 16-shot)} \\
    \midrule
    CLIP ViT-B/16 & 99.3 & \textbf{87.1} & 99.6 & \textbf{84.2} \textbar \textbf{99.7} \\
    DINOv2 ViT-B/14 & - & 81.8 & - & 76.2 \textbar 97.3\\
    DINOv3 ViT-L/16 & - & 84.3 & - & 80.2 \textbar 97.8 \\
    \bottomrule
\end{tabular}
}
\end{table}
However, we consider it unlikely that CLIP fails on such dataset.
A simple way to rule out a model weakness is comparison with uni-modal vision models such as DINO \cite{dinov2,dinov3}.
\cref{tab:catdog_retrieval} demonstrates how image-image metrics with DINO lack even more behind the text-image result, prompting the suggestion that the observed measures are \emph{not a weakness of CLIP}.
\pagebreak
\section{Methodology} % How you reassess
\label{sec:methodology}
Following the structure established by the previous supporting evidence from \cref{subsec:supporting}, our methodology to re-evaluate the intra-modal misalignment hypothesis is threefold:

From the theoretical perspective, \cref{subsec:dof} re-examines if text-image alignment leaves degrees of freedom for image-image misalignment. 

From the empirical perspective, \cref{subsec:nonclip} builds on the insight from the preliminary experiment
that we can substitute models to allow for controlled study of indicators.

Lastly, to explain performance gains of previous image-to-text augmentation approaches, \cref{subsec:simplealt} proposes a simple alternative method. % with image-image similarities only.

\subsection{Analyzing intra-modal degrees of freedom}
\label{subsec:dof}
\begin{figure}
    \centering
    \includegraphics[width=\linewidth]{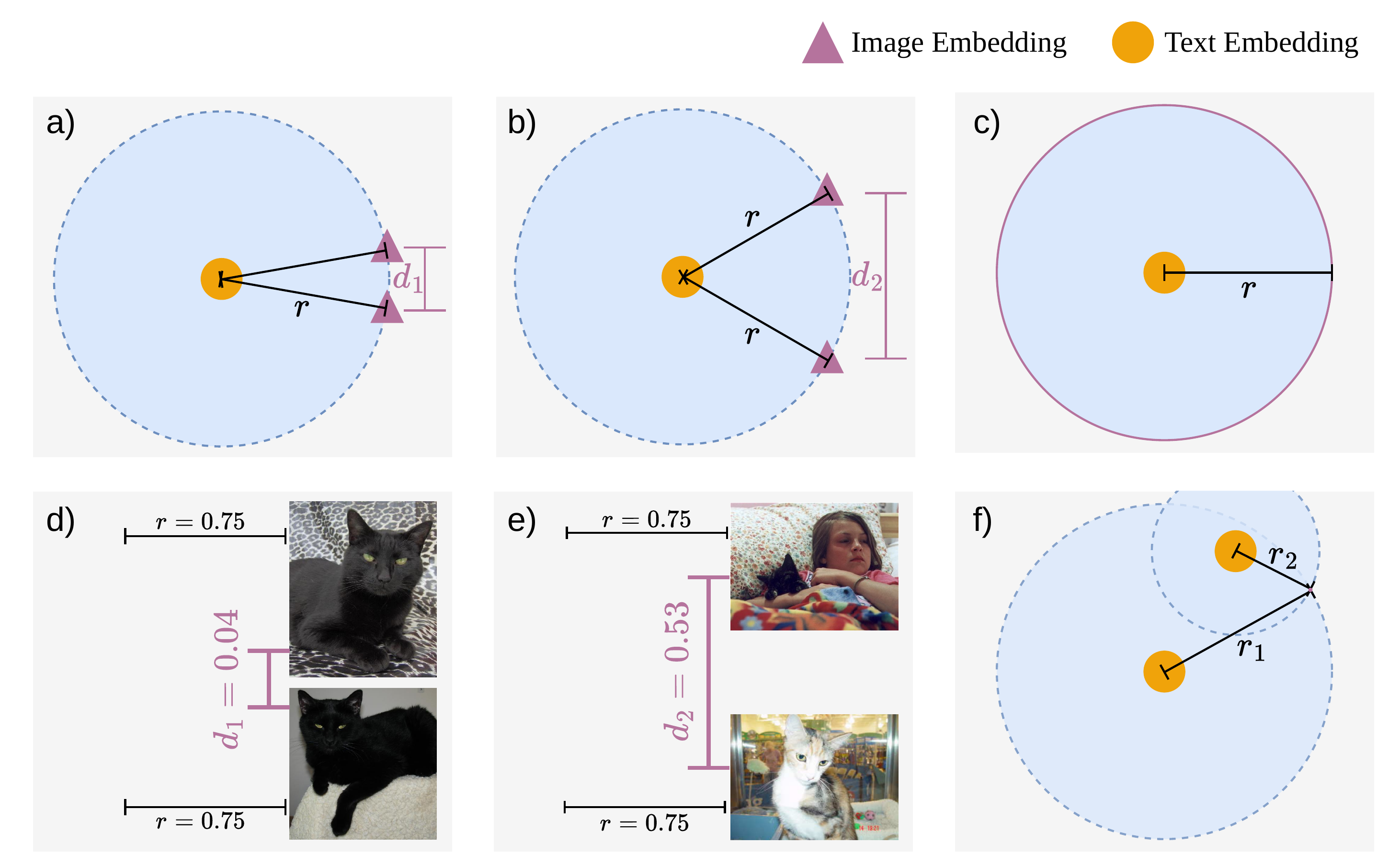}
    \caption{\textbf{(a)-(c): The previous intra-modal degree-of-freedom argument} in \cite{susx} illustrates that two image embeddings can be either close together (a) or far apart (b) while having the same text-image distance $r$, concluding that two image embeddings can lie on any two arbitrary points on the circumference (c), leaving a degree of freedom for image-image miscalibration.
    \textbf{Our interpretation (d)-(f):} The previous line of argumentation overlooks that each image embedding is bound to 
    more than one text anchor (f). 
    Moreover, the two different configurations in (a) and (b) are not arbitrary, but have a good reason to exist: Images in (a,d) and (b,e) have equal distance $r$ to the ``cat'' text, but the two images in (a,d) are much more similar to each other than those in (b,e). Displayed distance values are real measurements.} 
    \label{fig:dof}
\end{figure}

We begin with re-examining of the degree of freedom argument in \cite{susx}, where it was hypothesized that intra-modal misalignment emerges from the degrees of freedom that image embeddings have during contrastive language-image training.
The idea how this leaves room for poor calibration is illustrated in \cref{fig:dof}a-c.

We argue the shortcoming of this hypothesis and its visual demonstration lies in its reduction to a single text embedding.
If we extend their model to a larger number of embeddings, then the conclusion no longer holds.

The general acknowledged assumption is that text-image distances are well calibrated.
An argument for the misalignment states that given certain text-image distances, there is still room for miscalibrated image-image distances.
We can test this argument, formulating the question:

\begin{quote}
In the pre-training dataset with $n_T$ texts and $n_I$ images, given fixed pair-wise \textit{inter-}modal cosine similarities expressed through  matrix $S_{inter} \in \mathbb{R}^{n_T\times n_I}$,
is there any room for miscalibrated \textit{intra-}modal similarities $S_{intra} \in \mathbb{R}^{n_I\times n_I}$?% between image pairs?
\end{quote}

If there were a way to express $S_{intra}$ as an entangled consequence of $S_{inter}$, then we could conclude that there is no degree of freedom.
The underlying image embeddings $X_I \in \mathbb{R}^{n_I \times d}$ are the unknown.
Also the text embeddings $X_T \in \mathbb{R}^{n_T \times d}$ are in general not necessary to infer $S_{intra}$.
We can show that in a $d$-dimensional space, with only $d$ sampled text anchors represented by row indices $\mathcal{J}\subset\{1,\dots,n_T\}, |\mathcal{J}|=d$, we can set up a linear system that allows for unique recovery of all image embeddings $X_I \in \mathbb{R}^{n_I \times d}$. We know that:
\begin{equation}
    S_{inter}[\mathcal{J}] = X_T[\mathcal{J}] \cdot {X_I} ^\top.
\end{equation}
Solving for $X_I$, 
\begin{equation}
    \label{eq:reconstruct_V}
    X_I = ({X_T[J]})^{-1} \cdot {S_{inter}[J]}.
\end{equation}
Sampling at least $d$ text embeddings ensures $X_T[\mathcal{J}]$ is invertible, given that every row in $X_T$ is a unique embedding unit vector.
Typically $n_T,n_I \gg d$, so the number of required anchor points is comparatively small.
The logic behind \cref{eq:reconstruct_V} is illustrated for $d=2$ in \cref{fig:dof}f.
From \cref{eq:reconstruct_V} follows that all image-image similarities are \emph{well-defined without further degrees of freedom}:
\begin{equation}
    \label{eq:reconstruct_Sintra}
    S_{intra}=X_I {X_I}^\top.
\end{equation}
The dot product equals cosine similarity because the rows in $X_I$ are unit vectors.
\cref{eq:reconstruct_V} produces unit vectors provided that the rows of $X_T$ are normalized likewise.

While in this section we borrow the text anchor assumption from previous work for illustration, it should be noted that such fixed points do not exist in reality. Therefore, in Appx.\ \ref{sec:appdx_dof} we show an alternative way to infer $S_{intra}$ without sampled anchor embeddings.

\subsection{Contrasting with non-CLIP}
\label{subsec:nonclip}
We seek to put the previous proposed indicators for a misalignment under test.

To isolate effects specific to the lack of an intra-modal training objective, we employ the method of comparing models solely trained with inter-modal objectives (CLIP, SigLIP \cite{siglip}) with models trained with intra-modal objectives (DINO, SigLIP2 \cite{siglip2}).

As in \cite{ctg} and \cref{fig:cossimhist_prev}a, we measure same- vs. opposite class cosine similarity distributions for paired and unpaired samples.
Following \cite{susx, Barbier2025BridgingTM} and \cref{fig:cossimhist_prev}b, we further measure same- vs. opposite modality similarity ($S_{intra}$ vs $S_{intra}$) distributions by substituting CLIP models with models that have additional image supervision as in the DINO objective.
Finally, we conduct the same CLIP vs. non-CLIP comparison on performance metrics such as retrieval average precision and few-shot classification accuracy. 

This way, if an experimental outcome reproduces for both model types, it cannot be attributed to a missing intra-modal term in CLIP's loss function.

\subsection{A simple alternative for more ``classness'' in similarity measure}
\label{subsec:simplealt}
When relying on similarity measures between images, many typical classification and retrieval tasks would benefit from only focusing on the image's single most dominant concept.
Ignoring all visual details and background objects would then remove information that does not correlate with the dataset's labels.

We attribute previous improvements brought by the modality inversion \cite{originaloti} technique from \cref{fig:oti} to this effect.
This technique was used in \cite{ctg} as a means to overcome intra-modal misalignment by converting an image to a text token.
This text token ${v^*}$ within the ``a photo of a $v^*$'' prompt is tuned through backpropagation through the text encoder in order to approximate the image embedding.

We argue this is only efficient because it forces the information contained in the image to collapse to a single word-like token.
The resulting text embedding is like ``What single word describes the image best?'', which very often correlates strongly with the to-predict class in the test dataset.

To validate our claim, we propose to replace the entire Modality Inversion procedure through simply projecting the image embeddings on the subspace that explains the most variance of ``a photo of $x$'' prompts, where $x$ is a word-like class name.
This is similar to \cite{Zhu2025ProjectProbeAggregateEF, Zhu2025EnhancingCR}.
Irrespective of the downstream dataset, we choose ImageNet class names for $x$.
Given $n$ class names, we extract $n$ text embeddings with dimension $d$.
The first $d/2$ principal components are selected.
All test images are projected onto the subspace spanned by the selected components.
We then obtain results by measuring image-image cosine similarities in this subspace.
We refer to this method as $PCA^{\leftarrow}$ in the experiments.
\section{Results}
\label{sec:results}
\cref{subsec:cosim_res} evaluates the cosine similarity histogram indicators under the CLIP/non-CLIP substitution methodology from \cref{subsec:nonclip}.
\cref{subsec:fs_results} and \cref{subsec:retrieval_results} present results on few-shot classification and image-to-image retrieval, respectively, including the method from \cref{subsec:simplealt}.

\subsection{Datasets and Metrics}
Following the line of works \cite{coop, zhou2022cocoop, tipadapter} on few-shot classification with CLIP, we evaluate on 11 image classification datasets. These datasets encompass a variety of image recognition tasks, such as generic object recognition with ImageNet \cite{deng2009imagenet} and Caltech101 \cite{wah2011caltech}, fine-grained recognition using OxfordPets \cite{oxfordpets}, StanfordCars \cite{stanfordcars}, Flowers102 \cite{flower102}, Food101 \cite{bossard2014food}, and FGVCAircraft \cite{maji2013fine}, satellite image classification with EuroSAT \cite{helber2019eurosat}, action classification with UCF101 \cite{soomro2012ucf101}, texture classification with DTD \cite{Cimpoi2013DescribingTI}, and scene recognition with SUN397 \cite{xiao2010sun}.
We measure the common classification accuracy metric.

We use these datasets also for image-to-image retrieval, following the work that hypothesized an image-image misalignment \cite{ctg}.
For retrieval, we additionally evaluate on the more commonly used dataset ROxford and RParis \cite{radenovic2018revisiting}.
For retrieval evaluations, following \cite{ctg}, we measure mean average precision (mAP) over retrieval queries.

\begin{figure}[htbp]
    \centering
    % First row
    \begin{subfigure}{0.48\linewidth}
        \centering
        \includegraphics[width=\textwidth]{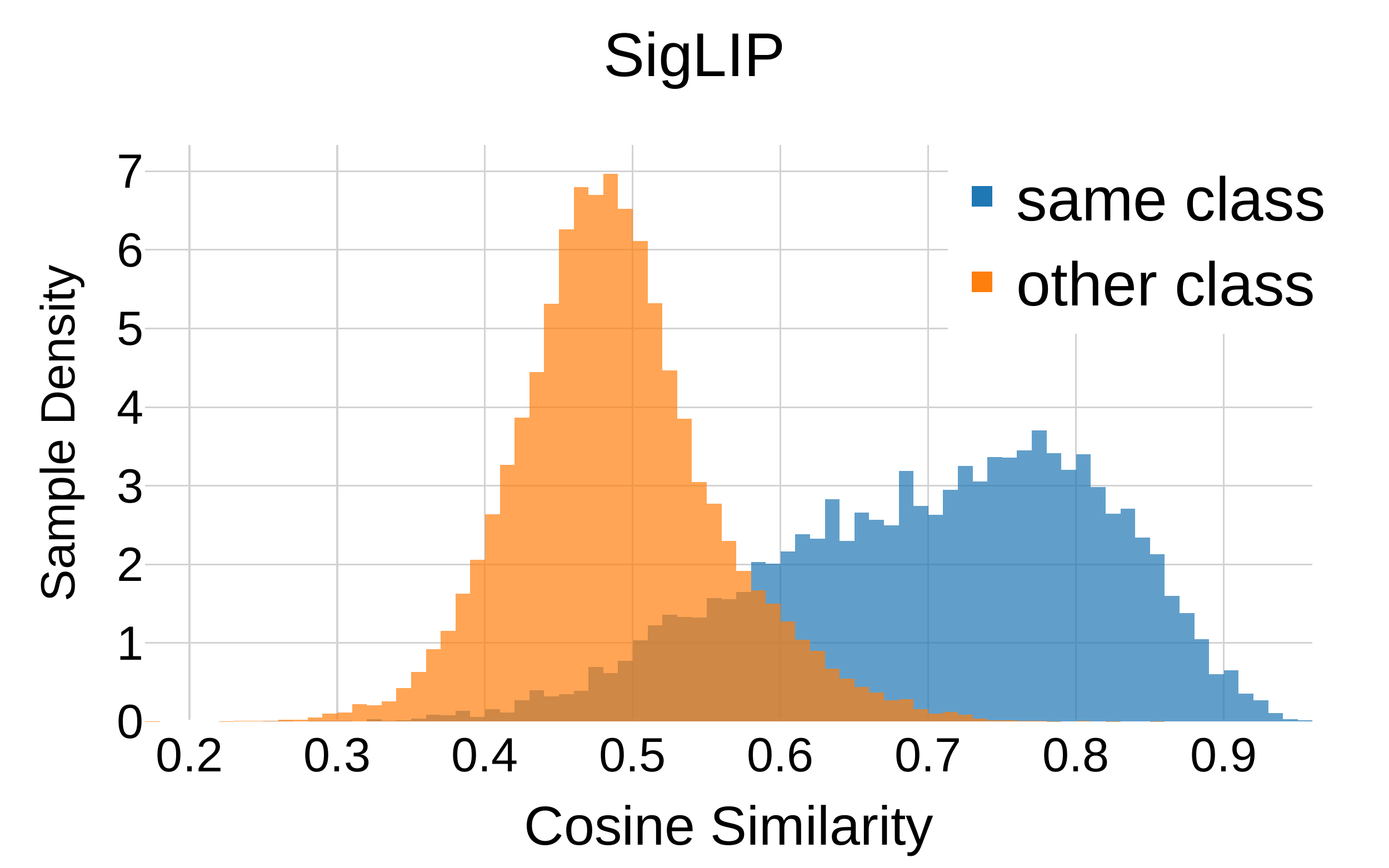}

    \end{subfigure}
    \hfill
    \begin{subfigure}{0.48\linewidth}
        \centering
        \includegraphics[width=\textwidth]{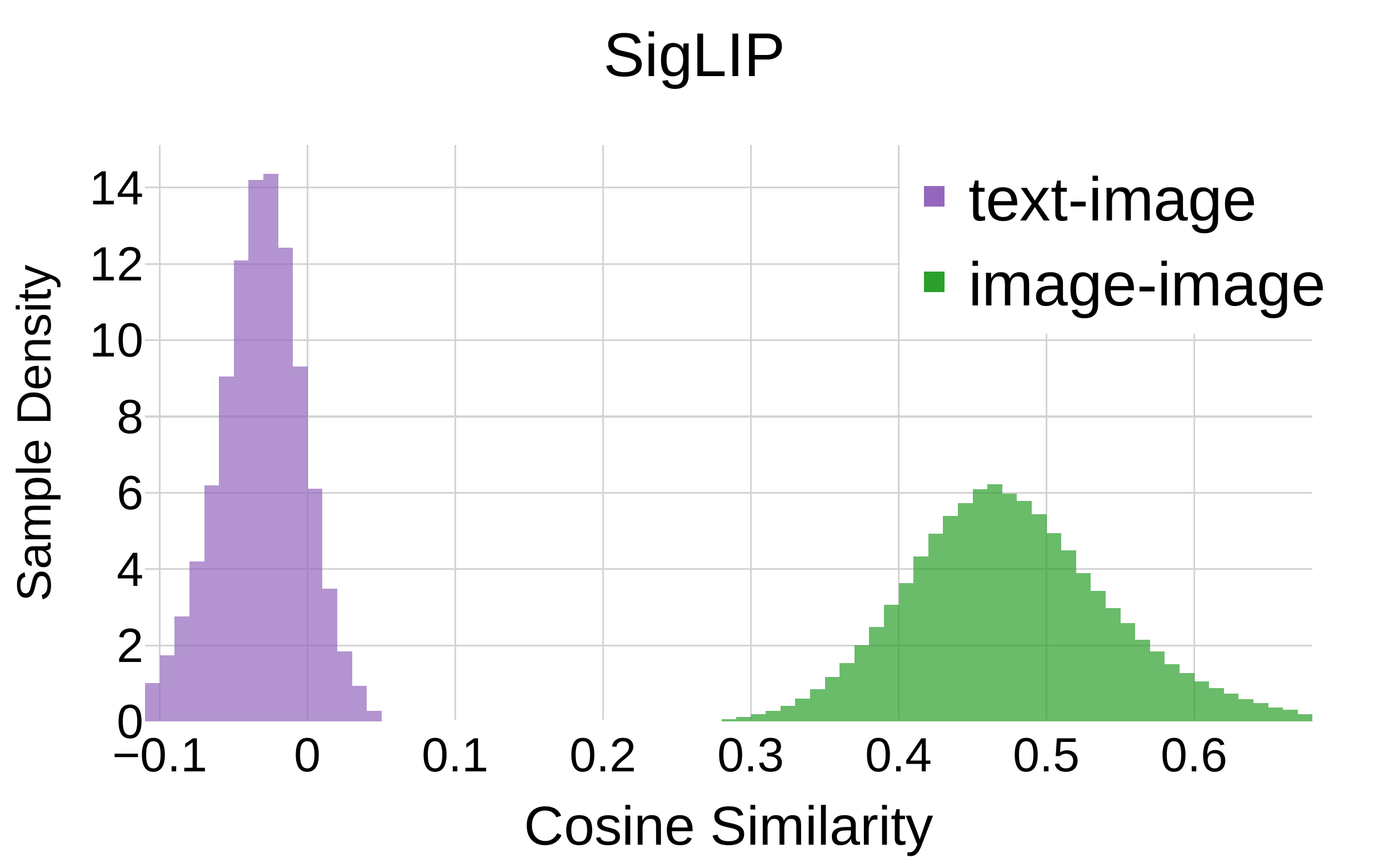}

    \end{subfigure}
    
    % Second row
    \begin{subfigure}{0.48\linewidth}
        \centering
        \includegraphics[width=\textwidth]{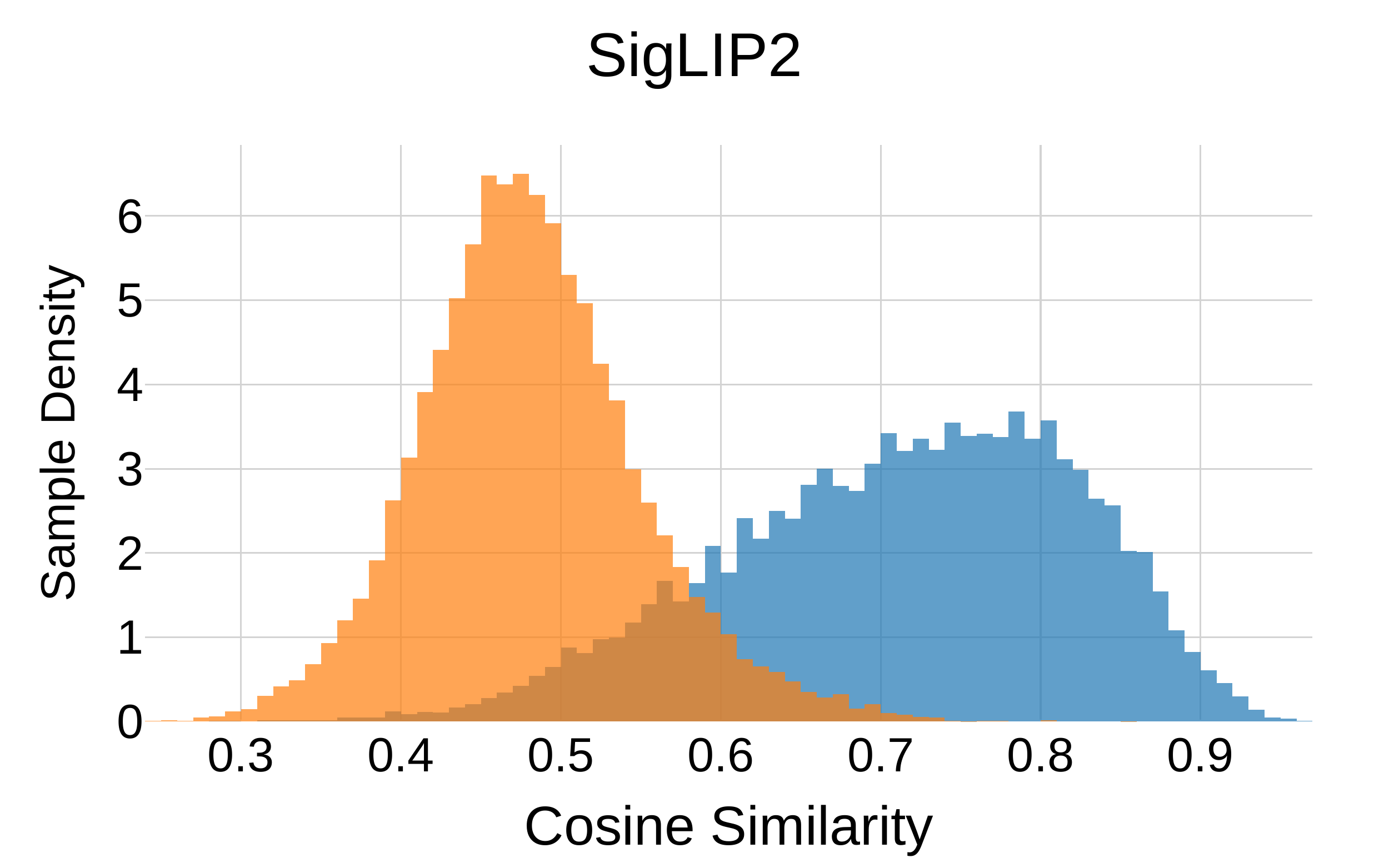}

    \end{subfigure}
    \hfill
    \begin{subfigure}{0.48\linewidth}
        \centering
        \includegraphics[width=\textwidth]{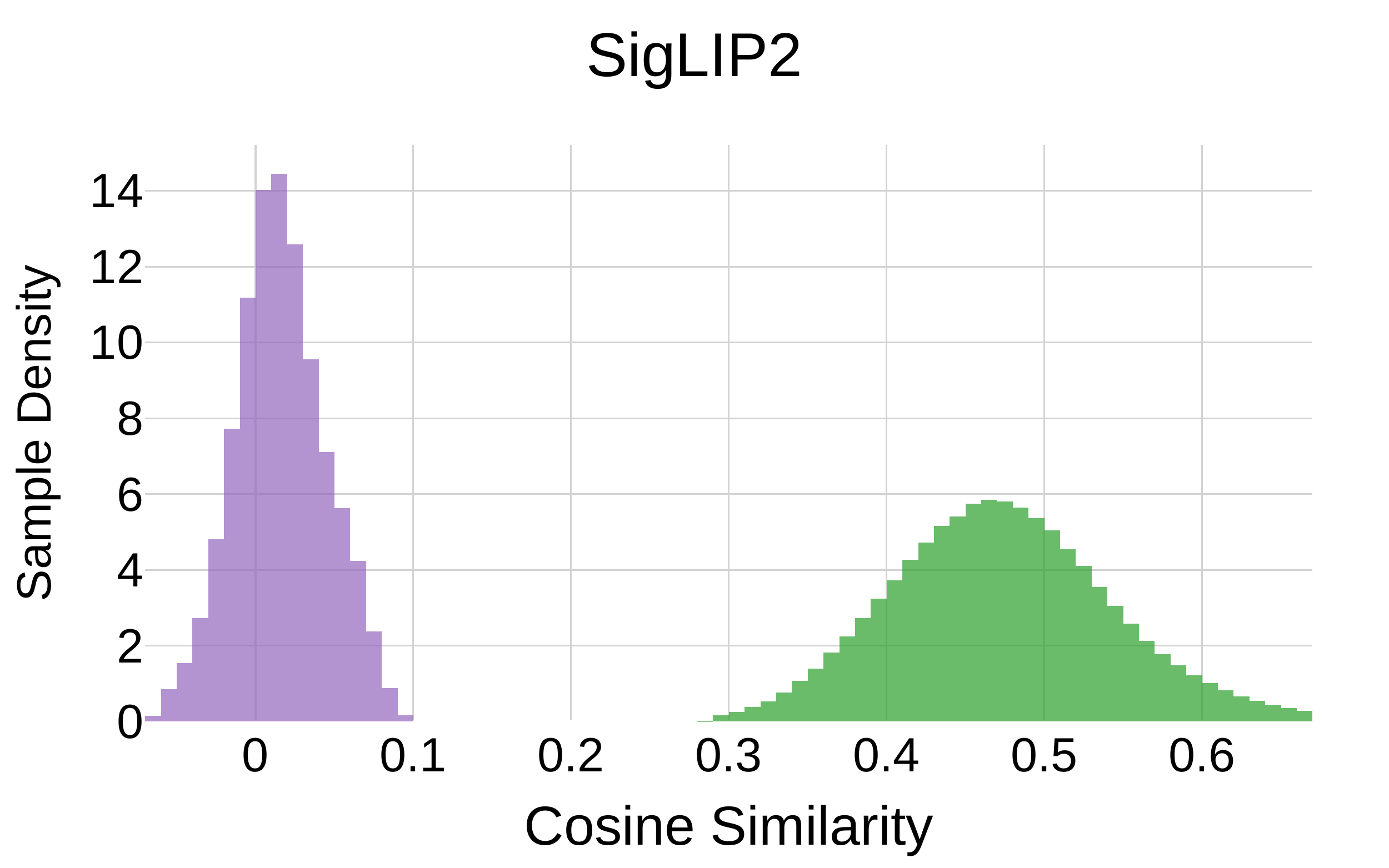}
        
    \end{subfigure}
    
    \caption{Cosine similarity histograms \textbf{by class (left)} and \textbf{by modality (right)}. The distributions are almost identical for purely text-image trained SigLIP (first row) and SigLIP2 (second row) which includes an image-image self-supervised objective as in the DINO line of work.
    This indicates intra-class variation (left) and the gap between text and image embeddings (right) are no signs of misalignment brought by pure text-image training, but rather normal behavior. All models are ViT-B. Embeddings are sampled from ImageNet validation set.}
    \label{fig:cossimhistos_res}
\end{figure}

\begin{figure*}[htbp]
    \centering
    \includegraphics[width=0.99\linewidth]{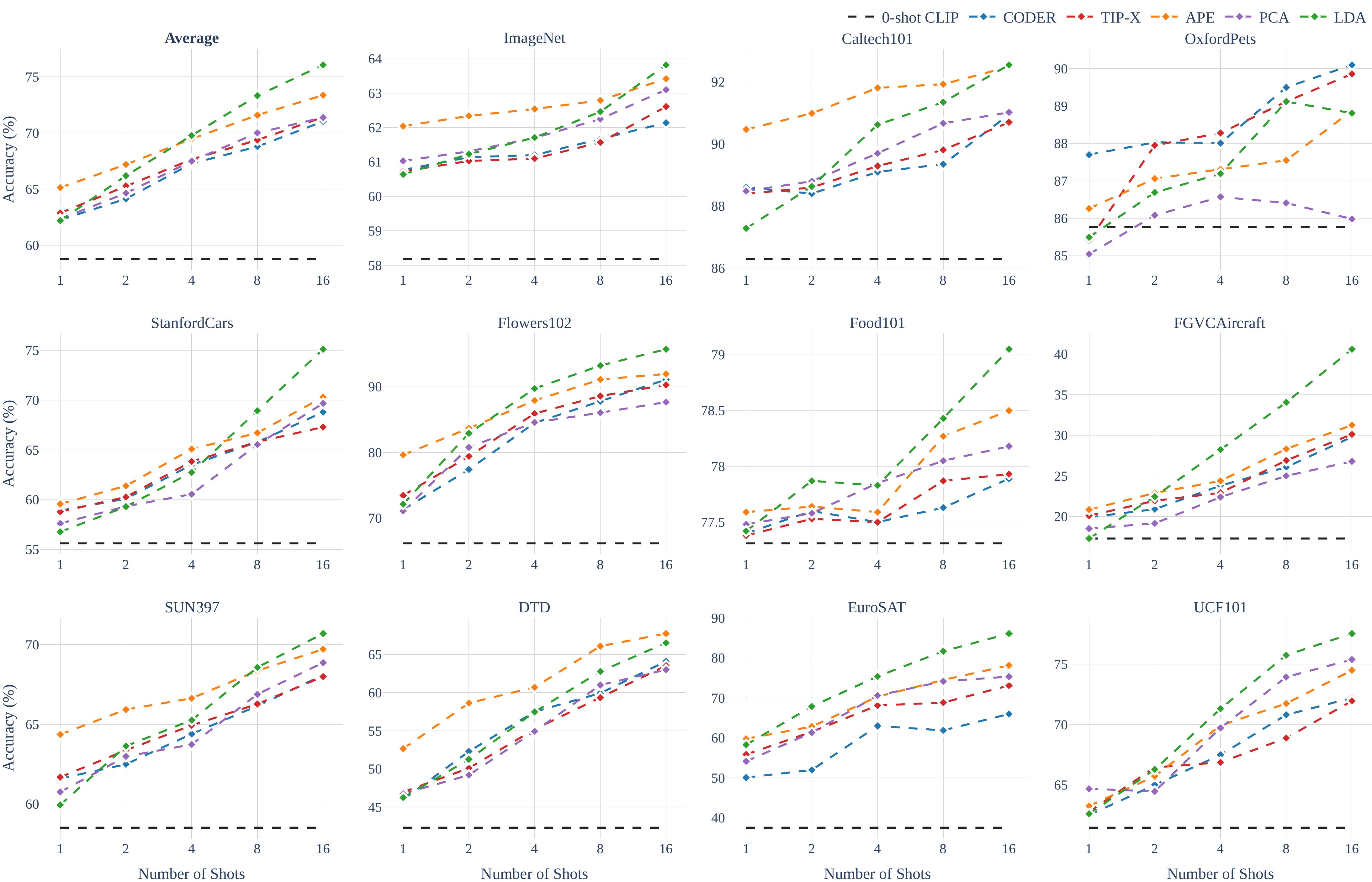}
    \caption{\textbf{Few-shot classification as in training-free VLM adaptation} literature. Approaches that attempt to address the intra-modal misalignment issue through avoiding use of image-image similarities (Tip-X, Coder) \cite{susx, coder} do not win against feature selection in APE \cite{ape} or a basic Linear Discriminant Analysis (LDA) \cite{Wang2024AHB} on image embeddings.}
    \label{fig:fsc_results}
\end{figure*}

\subsection{Cosine Similarity Distributions}
\label{subsec:cosim_res}
\textbf{Setup.} 
We analyze the two cosine similarity distribution indicators proposed as evidence for intra-modal misalignment (see \cref{fig:cossimhist_prev} for a recap).
To test whether these indicators are specific to models lacking an intra-modal objective, we contrast a model trained with only an inter-modal loss (SigLIP) with a model that includes an additional image-image objective (SigLIP2). If the indicators are caused by a missing intra-modal loss, they should be ``fixed'' in SigLIP2.
Results are shown in \cref{fig:cossimhistos_res}.
The key insight is that both indicators are indistinguishable between the two models, with details discussed below.

\textbf{Class distances.}
\cref{fig:cossimhistos_res} (left) shows that both SigLIP and SigLIP2 produce cosine similarity distributions where there are many image pairs from the same class less similar to each other than some image pairs from a different class, mirroring the introductory cat-cat cat-dog example in \cref{fig:introfig}.
The finding that also SigLIP2 manifests the high overlaps in the histograms further supports our alternative hypothesis that this is not a sign of misalignment,
but an expected and desirable property of general-purpose pretrained vision encoders.
Moreover, the result suggests that SigLIP2's improvement over SigLIP is not due to a better class separation that can be visualized through the used histogram.

\textbf{Modality gap.}
\cref{fig:cossimhistos_res} (right) leads to the same conclusion.
The divergence in similarity distributions between inter-modal (image-text) and intra-modal (image-image) pairs is nearly identical for both models.
The fact that SigLIP2's additional image-image training does not close this gap demonstrates that it is a misalignment introduced by a missing image-image objective.
We do not see that this divergence is harmful for intra-modal tasks.
In fact, the different range can be explained straightforward as a consequence of the modality gap:
Because image and text embeddings cluster in two distinct regions in the embedding space, distances measured within the cluster are naturally smaller than between clusters.
We discuss this in context of findings from prior work in \cref{sec:discussion}. 

\subsection{Comparison on the Few-Shot Classification Task}
\label{subsec:fs_results}
\textbf{Setup.}
We study few-shot classification performance in two complementary settings: 
(1) the \emph{standard VLM few-shot adaptation} setting \cite{coop,zhou2022cocoop} where a text prompt provides a zero-shot classifier that is refined using few-shot labeled images; and 
(2) an \emph{image-only few-shot} setting with no text prompts, designed to isolate the intrinsic alignment quality of the image embeddings. 
The former corresponds to \cref{fig:fsc_results}, the latter to \cref{tab:fsc_results_imgonly}.

\textbf{Few-Shot VLM Adaptation.}
Following prior work, CLIP ResNet-50 results are reported.
\cref{fig:fsc_results} summarizes performance across 11 datasets.
Methods such as TIP-X and Coder were explicitly designed to circumvent potential intra-modal misalignment by avoiding image--image similarity.  
However, the results show that these approaches do not outperform simple image-space discriminative baselines such as APE feature selection or Linear Discriminant Analysis (LDA). \footnote{While in the original work the method was introduced as Gaussian Discriminant Analysis (GDA), we here refer to it as Linear Discriminant Analysis (LDA) because they assume a shared identical covariance matrix for all classes.}
The reported $PCA^{\leftarrow}$ results simply use the class mean of the projected few-shot samples as classifier. It performs competitively with Tip-X and Coder, supporting the view that classification in the image space is sufficient, without requiring the avoidance of image–image similarity.

\textbf{ Image-Only Few-Shot Classification.}
We evaluate few-shot classification \emph{without} any text prompts, more akin to a classical uni-modal few-shot learning setup \cite{fsl}. 
Classification is performed measuring image-image cosine-similarity to prototypes (class means) or by estimating the classifier as in \cite{Wang2024AHB}.
The two classifiers are tested with original and $PCA^{\leftarrow}$ projected embeddings.
CLIP ViT-B/16 is used.

Across 11 datasets, \cref{tab:fsc_results_imgonly} evidences how
SigLIP (inter-modal only) outperforms DINOv2 (image-only) in image-image similarity based classification. 
This ranking is consistent across classifiers.  
While comparing models that have been trained on different data does not allow for conclusions about the impact of the training objective (loss function), the experiment still reveals a notable result relevant to the misalignment hypothesis:
pure contrastive language-image training can yield image embeddings that are aligned well enough to outperform strong pure vision encoders.
In other words, the results lack any evidence of misalignment.

\begin{table}[htbp]
\caption{\textbf{Few-shot classification with no use of text prompts.} 
Accuracy is therefore only based on image-to-image similarities. This allows directly to test the intra-modal alignment quality of models with \smash{\colorbox{trainedinter}{inter-modal}} loss only (CLIP, SigLIP), both inter- and \smash{\colorbox{trainedintra}{intra-modal}} losses (SigLIP2) and image-only models (DINO). Results on $PCA^{\leftarrow}$-projected features are in ($\cdot$) parenthesis. Two best models per classifier are \underline{underlined}.
\smash{\colorbox{trainedinter}{SigLIP}} scores higher than \smash{\colorbox{trainedintra}{DINO}}, indicating well-aligned image embeddings despite inter-modal training only. All models are ViT-B. Reported values are the average over the 11 datasets as in \cref{fig:fsc_results}.}
% For this experiment you cannot test inter-modal methods like Tip-X, because it needs the text embeddings.
\vspace{-5pt}
\label{tab:fsc_results_imgonly}
\centering
\resizebox{1.0\linewidth}{!}{
\begin{tabular}{lllllll} %ccccc
    \toprule
    \textbf{Model} & \textbf{Classifier} & \textbf{1-shot} & \textbf{2-shot} & \textbf{4-shot} & \textbf{8-shot} & \textbf{16-shot} \\
    \midrule
    \multirow{2}{*}{\shortstack{CLIP}}
    & Prototype & \cellcolor{trainedinter}{43.4 (50.7)} & \cellcolor{trainedinter}{55.3 (62.5)} & \cellcolor{trainedinter}{63.8 (69.7)} & \cellcolor{trainedinter}{69.6 (74.7)} & \cellcolor{trainedinter}{73.5 (77.5)} \\
    & LDA & \cellcolor{trainedinter}{43.4 (50.7)} & \cellcolor{trainedinter}{60.0 (62.5)} & \cellcolor{trainedinter}{69.8 (70.8)} & \cellcolor{trainedinter}{76.1 (76.5)} & \cellcolor{trainedinter}{79.5 (79.6)} \\
    \cmidrule{1-7}
    
    \multirow{2}{*}{\shortstack{SigLIP}}
    & Prototype & \cellcolor{trainedinter}{{57.3} (62.1)} & \cellcolor{trainedinter}{\underline{68.6} (71.9)} & \cellcolor{trainedinter}{\underline{76.3} (78.5)} & \cellcolor{trainedinter}{\underline{80.5} (81.8)} & \cellcolor{trainedinter}{\underline{82.5} (83.7)} \\
    & LDA & \cellcolor{trainedinter}{{57.3} (62.1)} & \cellcolor{trainedinter}{\underline{71.0} (72.4)} & \cellcolor{trainedinter}{\underline{79.0} (79.4)} & \cellcolor{trainedinter}{\underline{83.3} (83.3)} & \cellcolor{trainedinter}{\underline{85.3} (85.3)} \\
    \cmidrule{1-7}
    
    \multirow{2}{*}{\shortstack{SigLIP2}}
    & Prototype & \cellcolor{trainedintra}{\underline{58.0} (63.4)} & \cellcolor{trainedintra}{\underline{69.7} (73.2)} & \cellcolor{trainedintra}{\underline{77.0} (79.4)} & \cellcolor{trainedintra}{\underline{80.8} (82.6)} & \cellcolor{trainedintra}{\underline{83.0} (84.5)} \\
    & LDA & \cellcolor{trainedintra}{\underline{58.0} (63.4)} & \cellcolor{trainedintra}{\underline{73.2} (74.5)} & \cellcolor{trainedintra}{\underline{80.5} (81.0)} & \cellcolor{trainedintra}{\underline{84.5} (84.7)} & \cellcolor{trainedintra}{\underline{86.5} (86.5)} \\
    \cmidrule{1-7}
    
    \multirow{2}{*}{\shortstack{DINOv2}}
    & Prototype & \cellcolor{trainedintra}\underline{59.6} & \cellcolor{trainedintra}{67.1} & \cellcolor{trainedintra}{71.8} & \cellcolor{trainedintra}{76.0} & \cellcolor{trainedintra}{78.2} \\
    & LDA & \cellcolor{trainedintra}\underline{59.6} & \cellcolor{trainedintra}{69.2} & \cellcolor{trainedintra}{75.3} & \cellcolor{trainedintra}{80.3} & \cellcolor{trainedintra}{83.3} \\
    \bottomrule
\end{tabular}
}
\vspace{-10pt}
\end{table}

\begin{table*}[htbp]
\caption{
\textbf{Image-to-image retrieval} results (mAP) following \cite{ctg}. Rows with $\langle I,I\rangle$ or $\langle I^\leftarrow,I^\leftarrow\rangle$ measure intra-modal similarities. We show the performance gain brought by OTI \cite{ctg} - an image-to-text conversion attempt motivated by the hypothesis that only $\langle T,I\rangle$ is aligned - can be significantly surpassed through simply measuring $\langle I^\leftarrow,I^\leftarrow\rangle$ along the main axes of variance of ImageNet class names. This trend is no different between models  \colorbox{trainedinter}{without intra-modal} and \colorbox{trainedintra}{with intra-modal} training objectives.
}
\vspace{-5pt}
\label{tab:image_retrieval}
    \centering
    \resizebox{.975\linewidth}{!}{
    \begin{tabular}{ccccccccccccccccc|c}
        \toprule
         & ViT & Method & $\langle \cdot,\cdot\rangle$ & \rotatebox[origin=lb]{90}{\smash{$\mathcal{R}$Oxford}} & \rotatebox[origin=lb]{90}{\smash{$\mathcal{R}$Paris}} & \rotatebox[origin=lb]{90}{\smash{Cars}} & \rotatebox[origin=lb]{90}{\smash{Pets}} & \rotatebox[origin=lb]{90}{\smash{Flowers}} & \rotatebox[origin=lb]{90}{\smash{Aircraft}} & \rotatebox[origin=lb]{90}{\smash{DTD}} & \rotatebox[origin=lb]{90}{\smash{EuroSAT}} & \rotatebox[origin=lb]{90}{\smash{Food101}} & \rotatebox[origin=lb]{90}{\smash{SUN397}} & \rotatebox[origin=lb]{90}{\smash{Caltech}} & \rotatebox[origin=lb]{90}{\smash{UCF101}} & \rotatebox[origin=lb]{90}{\smash{ImageNet}} & \rotatebox[origin=lb]{90}{\smash{\textbf{Average}}} \\
        \midrule
        
    \multirow{6}{*}{\rotatebox[origin=c]{90}{CLIP}}
    & \multirow{3}{*}{B/32} & Original & $\langle I,I\rangle$ & \cellcolor{trainedinter}42.4 & \cellcolor{trainedinter}74.0 & \cellcolor{trainedinter}24.9 & \cellcolor{trainedinter}31.2 & \cellcolor{trainedinter}62.5 & \cellcolor{trainedinter}14.5 & \cellcolor{trainedinter}28.3 & \cellcolor{trainedinter}49.3 & \cellcolor{trainedinter}33.6 & \cellcolor{trainedinter}34.6 & \cellcolor{trainedinter}77.7 & \cellcolor{trainedinter}46.2 & \cellcolor{trainedinter}21.6 & \cellcolor{trainedinter}41.6 \\ 
    
    & & OTI & $\langle T,I\rangle$ & \cellcolor{trainedinter}{43.0} & \cellcolor{trainedinter}{70.3} & \cellcolor{trainedinter}{28.0} & \cellcolor{trainedinter}{37.5} & \cellcolor{trainedinter}{62.6} & \cellcolor{trainedinter}14.4 & \cellcolor{trainedinter}{31.9} & \cellcolor{trainedinter}47.2 & \cellcolor{trainedinter}{34.7} & \cellcolor{trainedinter}{36.3} & \cellcolor{trainedinter}{79.9} & \cellcolor{trainedinter}{48.6} & \cellcolor{trainedinter}{23.8} & \cellcolor{trainedinter}{42.9} \\
    
    & & $\text{PCA}^{\leftarrow}$  & $\langle I^\leftarrow,I^\leftarrow\rangle$ & \cellcolor{trainedinter}\textbf{51.4} & \cellcolor{trainedinter}\textbf{80.9} & \cellcolor{trainedinter}\textbf{34.6} & \cellcolor{trainedinter}\textbf{47.7} & \cellcolor{trainedinter}\textbf{70.3} & \cellcolor{trainedinter}\textbf{16.1} & \cellcolor{trainedinter}\textbf{34.0} & \cellcolor{trainedinter}\textbf{53.8} & \cellcolor{trainedinter}\textbf{43.0} & \cellcolor{trainedinter}\textbf{40.0} & \cellcolor{trainedinter}\textbf{83.3} & \cellcolor{trainedinter}\textbf{53.3} & \cellcolor{trainedinter}\textbf{28.6} & \cellcolor{trainedinter}\textbf{49.0} \\
    
    \cmidrule{2-18}
    
    & \multirow{3}{*}{L/14} & Original & $\langle I,I\rangle$ & \cellcolor{trainedinter}57.1 & \cellcolor{trainedinter}77.8 & \cellcolor{trainedinter}43.8 & \cellcolor{trainedinter}47.2 & \cellcolor{trainedinter}84.2 & \cellcolor{trainedinter}25.8 & \cellcolor{trainedinter}33.9 & \cellcolor{trainedinter}57.8 & \cellcolor{trainedinter}55.0 & \cellcolor{trainedinter}39.2 & \cellcolor{trainedinter}83.8 & \cellcolor{trainedinter}59.5 & \cellcolor{trainedinter}33.0 & \cellcolor{trainedinter}53.7 \\ 
    
    & & OTI & $\langle T,I\rangle$ & \cellcolor{trainedinter}{62.4} & \cellcolor{trainedinter}{77.1} & \cellcolor{trainedinter}{50.5} & \cellcolor{trainedinter}{56.0} & \cellcolor{trainedinter}{86.0} & \cellcolor{trainedinter}27.1 & \cellcolor{trainedinter}{37.7} & \cellcolor{trainedinter}56.3 & \cellcolor{trainedinter}{55.9} & \cellcolor{trainedinter}{43.5} & \cellcolor{trainedinter}{87.3} & \cellcolor{trainedinter}{62.8} & \cellcolor{trainedinter}{38.2} & \cellcolor{trainedinter}{57.0} \\
    
    & & $\text{PCA}^{\leftarrow}$  & $\langle I^\leftarrow,I^\leftarrow\rangle$ & \cellcolor{trainedinter}\textbf{64.5} & \cellcolor{trainedinter}\textbf{83.0} & \cellcolor{trainedinter}\textbf{57.2} & \cellcolor{trainedinter}\textbf{62.7} & \cellcolor{trainedinter}\textbf{88.9} & \cellcolor{trainedinter}\textbf{28.7} & \cellcolor{trainedinter}\textbf{39.9} & \cellcolor{trainedinter}\textbf{62.8} & \cellcolor{trainedinter}\textbf{64.4} & \cellcolor{trainedinter}\textbf{46.0} & \cellcolor{trainedinter}\textbf{89.5} & \cellcolor{trainedinter}\textbf{66.8} & \cellcolor{trainedinter}\textbf{42.2} & \cellcolor{trainedinter}\textbf{61.3} \\
    
    \midrule
    
    \multirow{3}{*}{{\rotatebox[origin=c]{90}{\shortstack{Sig\\LIP}}}}
    & \multirow{3}{*}{B/16} & Original & $\langle I,I\rangle$ & \cellcolor{trainedinter}{50.6} & \cellcolor{trainedinter}{73.1} & \cellcolor{trainedinter}{65.7} & \cellcolor{trainedinter}{56.4} & \cellcolor{trainedinter}{87.5} & \cellcolor{trainedinter}{37.9} & \cellcolor{trainedinter}{39.8} & \cellcolor{trainedinter}{53.3} & \cellcolor{trainedinter}{56.3} & \cellcolor{trainedinter}{42.8} & \cellcolor{trainedinter}{87.2} & \cellcolor{trainedinter}{56.9} & \cellcolor{trainedinter}{35.8} & \cellcolor{trainedinter}57.2 \\
    & & OTI & $\langle T,I\rangle$ & \cellcolor{trainedinter}{55.2} & \cellcolor{trainedinter}{79.1} & \cellcolor{trainedinter}{71.8} & \cellcolor{trainedinter}{64.2} & \cellcolor{trainedinter}{89.7} & \cellcolor{trainedinter}{37.6} & \cellcolor{trainedinter}{43.3} & \cellcolor{trainedinter}{52.9} & \cellcolor{trainedinter}{59.0} & \cellcolor{trainedinter}{43.6} & \cellcolor{trainedinter}{88.9} & \cellcolor{trainedinter}{54.9} & \cellcolor{trainedinter}{38.8} & \cellcolor{trainedinter}{60.0} \\
    & & $\text{PCA}^{\leftarrow}$ & $\langle I^\leftarrow,I^\leftarrow\rangle$ & \cellcolor{trainedinter}\textbf{57.9} & \cellcolor{trainedinter}\textbf{78.4} & \cellcolor{trainedinter}\textbf{77.2} & \cellcolor{trainedinter}\textbf{68.5} & \cellcolor{trainedinter}\textbf{92.0} & \cellcolor{trainedinter}\textbf{40.9} & \cellcolor{trainedinter}\textbf{44.2} & \cellcolor{trainedinter}\textbf{54.2} & \cellcolor{trainedinter}\textbf{61.8} & \cellcolor{trainedinter}\textbf{46.9} & \cellcolor{trainedinter}\textbf{91.2} & \cellcolor{trainedinter}\textbf{60.3} & \cellcolor{trainedinter}\textbf{43.5} & \cellcolor{trainedinter}\textbf{62.8} \\
    \midrule
    
    \multirow{3}{*}{{\rotatebox[origin=c]{90}{\shortstack{SLIP}}}}
    & \multirow{3}{*}{B/16} &Original & $\langle I,I\rangle$ & \cellcolor{trainedintra}{36.5} & \cellcolor{trainedintra}{79.2} & \cellcolor{trainedintra}{4.9} & \cellcolor{trainedintra}{17.8} & \cellcolor{trainedintra}{65.7} & \cellcolor{trainedintra}{9.1} & \cellcolor{trainedintra}{29.7} & \cellcolor{trainedintra}{53.7} & \cellcolor{trainedintra}{19.5} & \cellcolor{trainedintra}{26.1} & \cellcolor{trainedintra}{65.4} & \cellcolor{trainedintra}{40.2} & \cellcolor{trainedintra}{15.4} & \cellcolor{trainedintra}35.6 \\ 
    & & OTI & $\langle T,I\rangle$ & \cellcolor{trainedintra}{36.4} & \cellcolor{trainedintra}{79.3} & \cellcolor{trainedintra}{5.0} & \cellcolor{trainedintra}{19.3} & \cellcolor{trainedintra}{65.1} & \cellcolor{trainedintra}{9.0} & \cellcolor{trainedintra}{30.5} & \cellcolor{trainedintra}{50.6} & \cellcolor{trainedintra}{20.0} & \cellcolor{trainedintra}{26.4} & \cellcolor{trainedintra}{67.6} & \cellcolor{trainedintra}{40.6} & \cellcolor{trainedintra}{14.8} & \cellcolor{trainedintra}{35.7} \\
    & & $\text{PCA}^{\leftarrow}$  & $\langle I^\leftarrow,I^\leftarrow\rangle$ & \cellcolor{trainedintra}\textbf{43.6} & \cellcolor{trainedintra}\textbf{83.9} & \cellcolor{trainedintra}\textbf{6.2} & \cellcolor{trainedintra}\textbf{23.1} & \cellcolor{trainedintra}\textbf{72.5} & \cellcolor{trainedintra}\textbf{10.3} & \cellcolor{trainedintra}\textbf{32.2} & \cellcolor{trainedintra}\textbf{53.2} & \cellcolor{trainedintra}\textbf{25.1} & \cellcolor{trainedintra}\textbf{30.7} & \cellcolor{trainedintra}\textbf{70.7} & \cellcolor{trainedintra}\textbf{42.4} & \cellcolor{trainedintra}\textbf{19.1} & \cellcolor{trainedintra}\textbf{39.4} \\ 
    \midrule
    
    \multirow{2}{*}{{\rotatebox[origin=c]{90}{\shortstack{Sig\\LIP2}}}}
    & \multirow{2}{*}{B/16} & Original & $\langle I,I\rangle$ & \cellcolor{trainedintra}52.5 & \cellcolor{trainedintra}75.6 & \cellcolor{trainedintra}{70.8} & \cellcolor{trainedintra}{56.6} & \cellcolor{trainedintra}{89.3} & \cellcolor{trainedintra}{40.7} & \cellcolor{trainedintra}{38.6} & \cellcolor{trainedintra}{49.3} & \cellcolor{trainedintra}{59.7} & \cellcolor{trainedintra}{43.0} & \cellcolor{trainedintra}{89.2} & \cellcolor{trainedintra}{59.2} & \cellcolor{trainedintra}{37.9} & \cellcolor{trainedintra}58.6 \\
    & & $\text{PCA}^{\leftarrow}$  & $\langle I^\leftarrow,I^\leftarrow\rangle$ & \cellcolor{trainedintra}\textbf{59.4} & \cellcolor{trainedintra}\textbf{78.6} & \cellcolor{trainedintra}\textbf{80.2} & \cellcolor{trainedintra}\textbf{67.8} & \cellcolor{trainedintra}\textbf{93.2} & \cellcolor{trainedintra}\textbf{46.3} & \cellcolor{trainedintra}\textbf{44.1} & \cellcolor{trainedintra}\textbf{51.1} & \cellcolor{trainedintra}\textbf{65.3} & \cellcolor{trainedintra}\textbf{48.9} & \cellcolor{trainedintra}\textbf{93.0} & \cellcolor{trainedintra}\textbf{63.0} & \cellcolor{trainedintra}\textbf{46.5} & \cellcolor{trainedintra}\textbf{64.4} \\ 
    \midrule
    
    \multicolumn{3}{c}{\rotatebox[origin=c]{0}{\shortstack{\textcolor{gray}{DINOv3 L/16}}}} & \textcolor{gray}{$\langle I,I\rangle$} & \cellcolor{trainedintra}{\textcolor{gray}{91.4}} & \cellcolor{trainedintra}{\textcolor{gray}{93.5}} & \cellcolor{trainedintra}{\textcolor{gray}{68.6}} & \cellcolor{trainedintra}{\textcolor{gray}{82.3}} & \cellcolor{trainedintra}{\textcolor{gray}{99.4}} & \cellcolor{trainedintra}{\textcolor{gray}{39.8}} & \cellcolor{trainedintra}{\textcolor{gray}{44.9}} & \cellcolor{trainedintra}{\textcolor{gray}{59.9}} & \cellcolor{trainedintra}{\textcolor{gray}{71.0}} & \cellcolor{trainedintra}{\textcolor{gray}{50.1}} & \cellcolor{trainedintra}{\textcolor{gray}{90.7}} & \cellcolor{trainedintra}{\textcolor{gray}{70.6}} & \cellcolor{trainedintra}{\textcolor{gray}{57.6}} & \cellcolor{trainedintra}{\textcolor{gray}{70.8}} \\
    \bottomrule
    \end{tabular}
    }
    %\vspace{-10pt}
\end{table*} 
\subsection{Comparison on the Retrieval Task}
\label{subsec:retrieval_results}

\textbf{Setup.}
We compare six models that we classify in inter-modal and intra-modal, following our contrasting approach.
Inter-modal models are those with no image self-supervised objective: CLIP B/32, L/14 and SigLIP.
Intra-modal models are SLIP \cite{slip}, SigLIP2 and DINO, that all have some self-supervised loss on the images.
CLIP, SigLIP and SLIP were also used in the OTI work \cite{ctg}, SigLIP2 and DINO were added by us in order to cover the intra-modal models.

\noindent \textbf{Results.}
\cref{tab:image_retrieval} reports our image-to-image retrieval results.
Our proposed $PCA^{\leftarrow}$ alternative consistently outperforms OTI across all datasets and across all models.

We consider this critical as it proves two key points our paper makes.
First, it shows comparing image embeddings works well, such that there is no reason to assume a misalignment and thus no need to invert images to pseudo-texts as in OTI and \cref{fig:oti}.
Second, it reveals no experimental outcome is specific to pure text-image training as in CLIP.
Specifically, the original models SigLIP and SigLIP2 perform comparable, with only $1.2$ mAP gain on SigLIP2, which is an expected improvement given it is an evolution of SigLIP, but no large gap which would point towards a significant problem in SigLIP with pure text-image training.
Also the degree to which the $PCA^{\leftarrow}$ projection helps is comparable, with $5.6$ and $5.8$ mAP gains, respectively.

We include DINOv3 L/16 as a reference upper limit.
Interestingly, it does not always score highest.
On Stanford Cars, it scores $8.6$ mAP lower than our SigLIP B/16 $PCA^{\leftarrow}$ from the inter-modal category.
On the other hand, DINO excels on the more standard retrieval datasets ROxford and RParis.
We attribute this to dataset curation: ROxford/RParis are designed for unambiguous retrieval, while the other datasets (to our knowledge only used in \cite{ctg}) suffer from query ambiguity akin to one-shot classification.
Their retrieval mAPs we show here are relevant for the study of the misalignment hypothesis, but for the task ambiguity reason, we  caution against viewing them as generally meaningful retrieval benchmarks in future work. 

\noindent \textbf{Ablation.}
To test our explanation why $PCA^{\leftarrow}$ can improve performance metrics, we conduct an experiment on a task where we assume it is \textit{not} helpful to focus on the dominant concept in the image.
Classifying the weather and time of the day requires an image representation that captures more than the main class, as background information matters.
We therefore measure retrieval metrics on ``weather'' and ``time of day'' classification on the BDD100k \cite{bdd100k} driving dataset.
Unlike on all datasets in \cref{tab:image_retrieval}, we find a performance \textit{drop} by $2.3\%$ mAP on weather and $1.0\%$ mAP on time of day compared to original CLIP (ViT-L/14).
The information loss through the projection is harmful here.
This confirms that $PCA^{\leftarrow}$ only brings gains if dataset labels align with the images' main semantic concept (\cref{fig:introfig}), but does not fix some misalignment in general.
\pagebreak
\section{Discussion}
\label{sec:discussion}
\textbf{Is the modality gap actually a problem?}
Given that previous papers titled ``Mind the gap'' \cite{Liang2022MindTG}, ``Mitigate the Gap'' \cite{Eslami2024MitigateTG}, ``Cross the gap'' \cite{ctg}, ``Bridging the Gap'' \cite{Barbier2025BridgingTM}, the gap may appear as a problem.
The resulting different distributions of cosine similarities (\cref{fig:cossimhist_prev}b, \cref{fig:cossimhistos_res}b) further made some \cite{susx,Barbier2025BridgingTM} belief something is intra-modally misaligned.
However, we failed to identify any such misalignment that negatively impacts image-only tasks.
Attempts of previous work to narrow the gap could also not bring consistent improvement \cite{Liang2022MindTG,ctg,Jiang2023UnderstandingAC}.
Eslami \etal \cite{Eslami2024MitigateTG} suggest adding an intra-modal separation loss reduces the gap while helpful for downstream performance, but this result is confined to their from-scratch trained modified shared text-vision encoder architecture.
Qian \etal \cite{Qian2023IntraModalPL} theoretically demonstrate that the gap cannot be reduced sufficiently by minimizing the contrastive loss in CLIP.
Jiang \etal \cite{Jiang2023UnderstandingAC} further prove that exact modality alignment is suboptimal in general for downstream prediction tasks.
We therefore see it as reasonable, not problematic, that image and text embeddings lie in two separate manifolds. %\todo{also "The Double-Ellipsoid Geometry of CLIP, ICML25"}
\\
\textbf{What about other tasks?} E.g.\ segmentation, depth estimation, VQA? --
We only evaluate on retrieval and few-shot classification because these were the tasks that were previously studied in the context of the intra-modal misalignment hypothesis.
To our knowledge, literature on other image-image applications of CLIP has not mentioned the misalignment issue.
If supposed misalignment turns out to be a non-issue for the previously studied tasks, we do not expect other tasks to be impacted, either.
As for the $PCA^\leftarrow$ method, we do not believe it will be useful for these tasks.
\\
\textbf{Did we disprove the hypothesis?}
For the experimental results, strictly speaking, no.
\cref{tab:catdog_retrieval,tab:fsc_results_imgonly,tab:image_retrieval,fig:cossimhistos_res} prove intra-modal misalignment was not the cause of the observed trends, but do not prove 
% intra-modal misalignment 
it is impossible to exist.
Similar to \cite{Schaeffer2023AreEA}, the refutation logic is to show that what previously was believed to be evidence for a hypothesis, can actually not serve as such.
For the theoretical argument in \cref{sec:methodology} and Appx.\ \ref{sec:appdx_dof}, yes it seeks to refute the belief in the underlying cause.
But ultimately, the hypotheses in prior work vary to some extent, so we believe it is more useful and appropriate to consider the evidence for the individual arguments than to speak of holistic proof/disproof.
\section{Conclusion}
In this work, we reevaluate the intra-modal misalignment hypothesis in CLIP-style vision-language models.
Through theoretical analysis, we demonstrate that intra-modal similarities are not unconstrained with arbitrary degrees of freedom, but determined by cross-modal similarities.
Empirically, we reexamine previously proposed indicators and performance metrics, showing they are no reliable diagnostic of intra-modal misalignment.
Comparison with a PCA-style projection method further supports our alternative hypothesis that task ambiguity in the few-shot setting, not misalignment, best explains the results.
We hope this can guide future work to leverage the pretrained intra-modal geometry rather than circumvent it.

\noindent \textbf{Acknowledgment}
{\small This work was supported by the National Nature Science Foundation of China under Grant No.\ 62522317 and 62373322, and by Zhejiang Provincial Natural Science Foundation of China under Grant No.\ LD25F030001.}
{
    \small
    \bibliographystyle{ieeenat_fullname}
    \bibliography{main}
}
\appendix
\clearpage
\setcounter{page}{1}
\maketitlesupplementary
\normalsize
\section{On degrees of freedom}
\label{sec:appdx_dof}
In Sec.\ 3.1 of the main paper, we presented a way to recover intra-modal similarities from inter-modal similarities.
This solution assumed a set of text anchors for simplification.
Here we show that even without this assumption, unique recovery of image-image similarities is possible:

\noindent \textbf{Start} with given inter-modal similarities
\begin{equation}
    S_{inter} = X_T  {X_I} ^\top,
\end{equation}
where the hidden underlying $X_T, X_I \in \mathbb{R}^{N \times d}$ have $d$-dim row vectors normalized to length 1 and $N\gg d$.

\noindent \textbf{Decompose} the $n \times n$ matrix $S_{inter}$ via SVD:
\begin{equation}
S_{inter} =  X_T  {X_I} ^\top = U \Sigma V^\top
\end{equation}
where:
\begin{itemize}
    \item $ \Sigma \in \mathbb{R}^{d \times d} $ is a diagonal matrix, we do not further use it.
    \item $ U, V \in \mathbb{R}^{N \times d }$ are orthogonal s.t. $ U^\top U = I = V^\top V $.
\end{itemize}

\noindent Since the columns of $ V $ span the same space as the columns of (full-rank) $ X_I $, we can write:
% Why?: Because S_{inter} was constructed from X_T X_I^T. Hence the column space of S_inter equals the column space of X_T
\begin{equation}
    X_I = V C
\end{equation}
for some $ d \times d $ matrix $ C $. Then:
\begin{equation}
\label{eq:vccv}
S_{intra} = X_I X_I^\top = V C C^\top V^\top.
\end{equation}
Let $ Q = C C^\top $. Because of normalized $X_I$, we know:
\begin{equation}
   \text{diag}(X_I X_I^\top) = \text{diag}(V Q V^T) = \mathbf{1}_N .
\end{equation}
 This gives a linear system for the entries of $ Q \in \mathbb{R}^{d \times d}$.
Specifically, there are $N$ quadratic forms
\begin{equation}
    v_i^\top Q v_i = \sum_{j=1}^{d} \sum_{k=1}^{d} V_{ij} V_{ik} Q_{jk} = 1 \quad \text{for } i = 1,...,N.
\end{equation}
\textbf{Solve} by rearranging in a standard $A x = b$ system,
\begin{itemize}[leftmargin=18pt]
    \item[$A:$] set the scalar product $V_{ij} V_{ik}$ as the entry of the $i$th row and $(jd-d+k)$th column in coefficient matrix $A \in \mathbb{R}^{N \times d^2}$.
    \item[$x:$] set flatten($Q$) as the variable vector $x \in \mathbb{R}^{d^2}$.
    \item[$b:$] set $b= \mathbf{1}_N \in \mathbb{R}^N$ .
\end{itemize}
There are $d (d+1)/2$ unknowns in $x$ since $Q$ is a symmetric $ d \times d $ matrix.
Since $ N \gg d $, the linear system $A x = b$ is overdetermined such that there is at most one solution for $x$ and there are \textbf{no degrees of freedom}.
In our recovery case, the solution exists; $b=1_N$ is in the column space of $ A $.
Solving for $x$, then reshaping $x$ back to $Q$, the intra-modal image-image similarities can be obtained via \cref{eq:vccv}:

\begin{equation}
S_{intra} = V Q V^T. 
\end{equation}
We provide a demonstration in PyTorch.

\begin{figure*}[htbp]
    \centering
    \begin{subfigure}{.49 \linewidth}
        \centering
        \includegraphics[width=\textwidth]{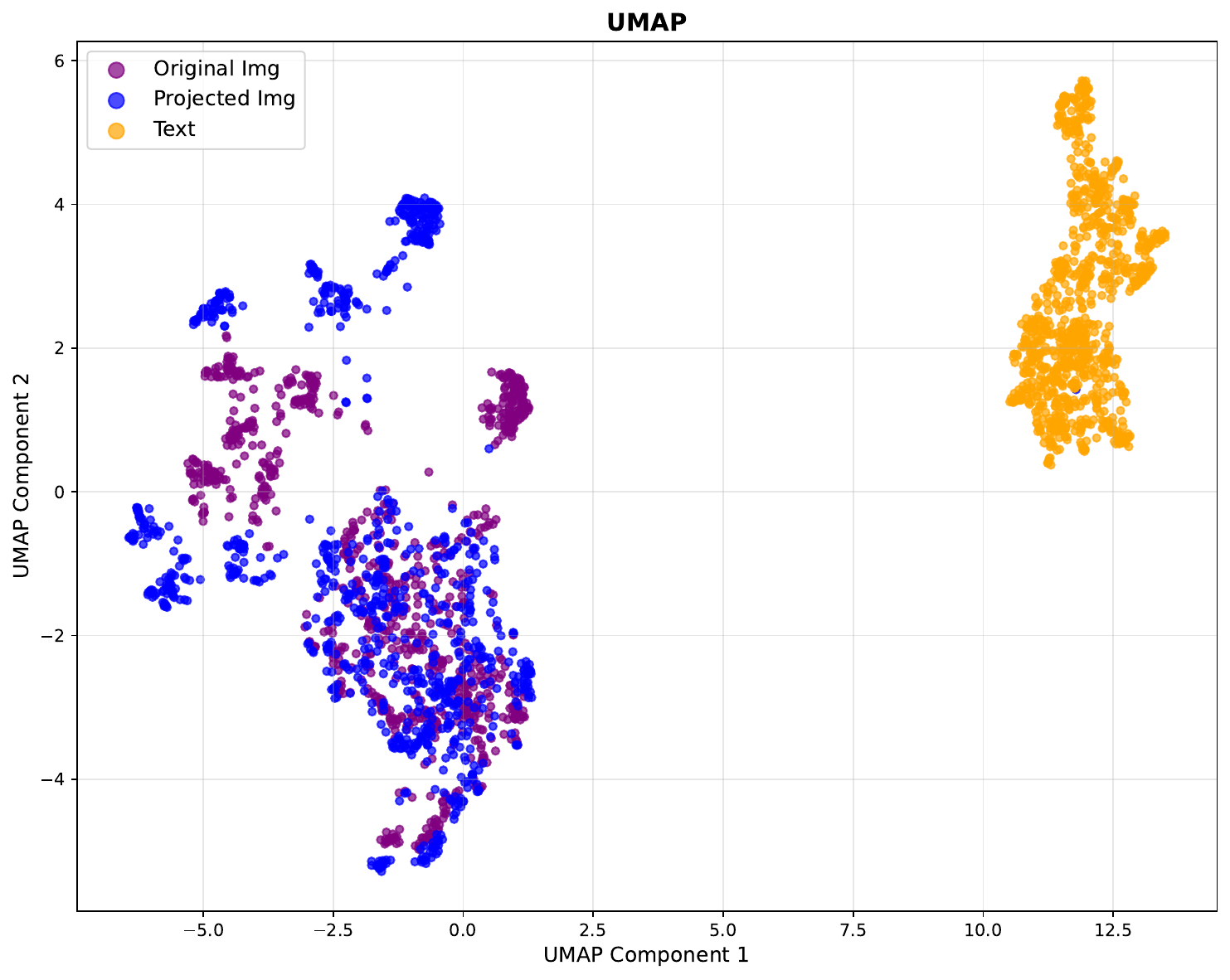}
    \end{subfigure}
    \begin{subfigure}{.49 \linewidth}
        \centering
        \includegraphics[width=\textwidth]{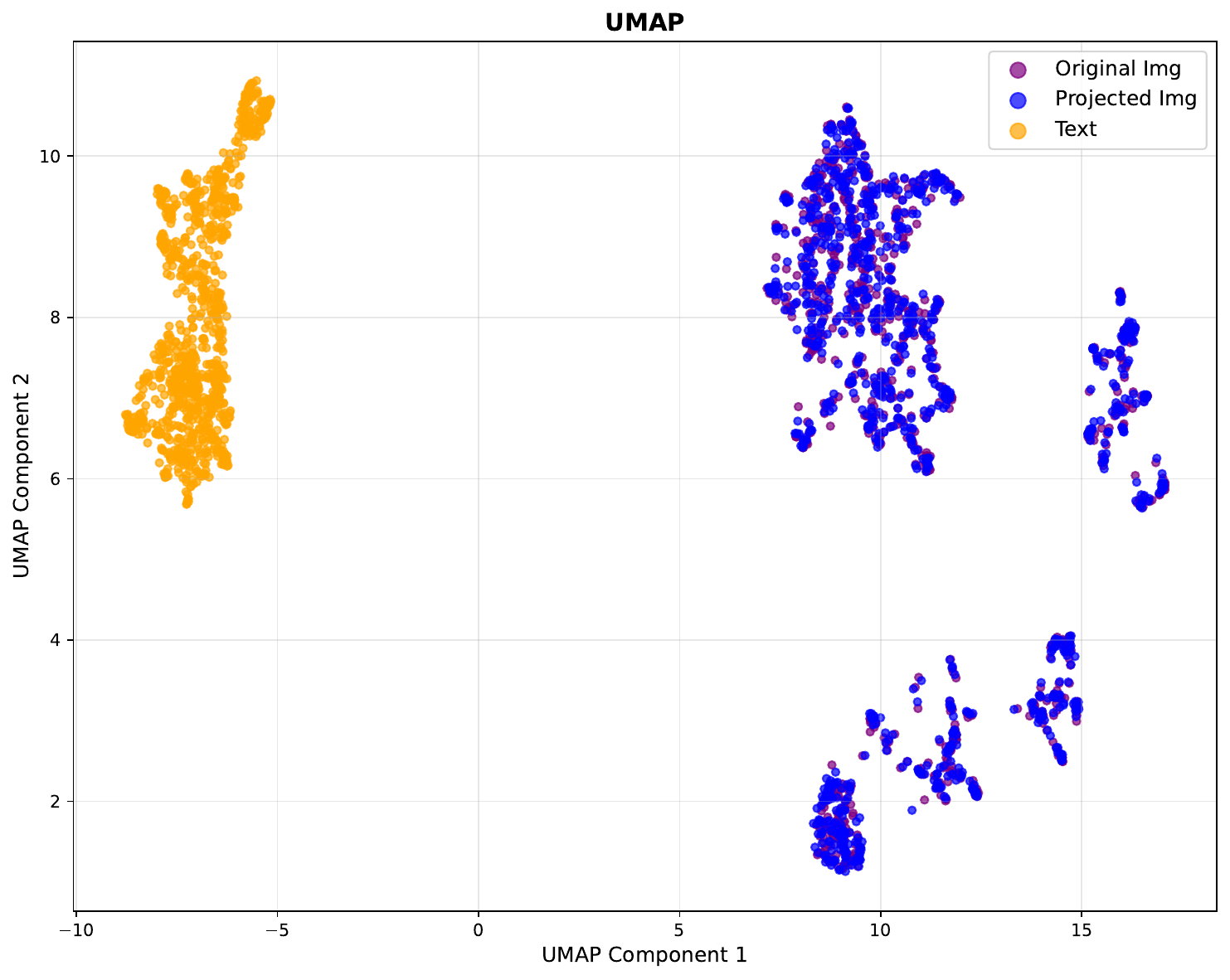}
    \end{subfigure}
    \hfill
     \begin{subfigure}{.49 \linewidth}
        \centering
        \includegraphics[width=\textwidth]{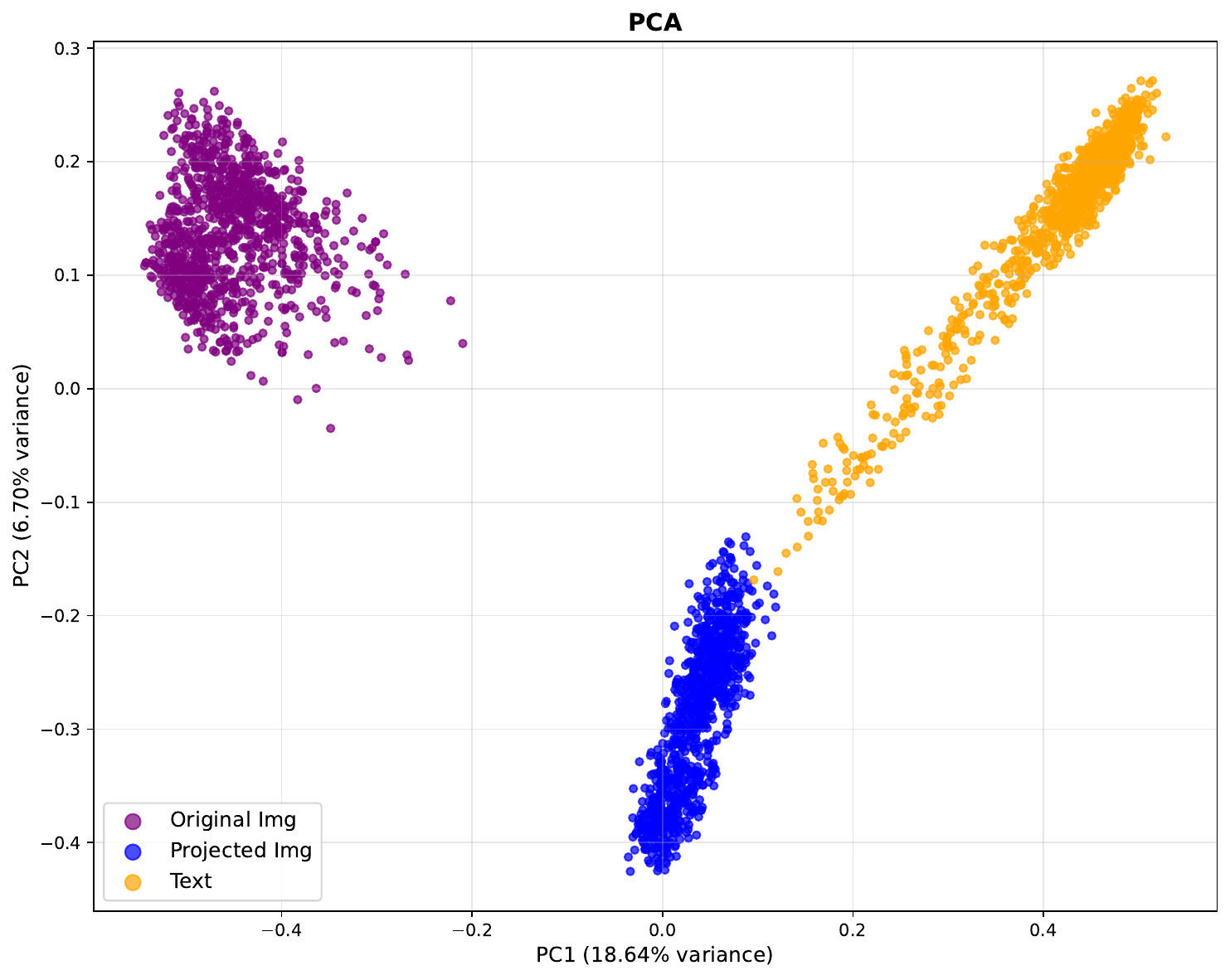}
    \end{subfigure}
    \begin{subfigure}{.49 \linewidth}
        \centering
        \includegraphics[width=\textwidth]{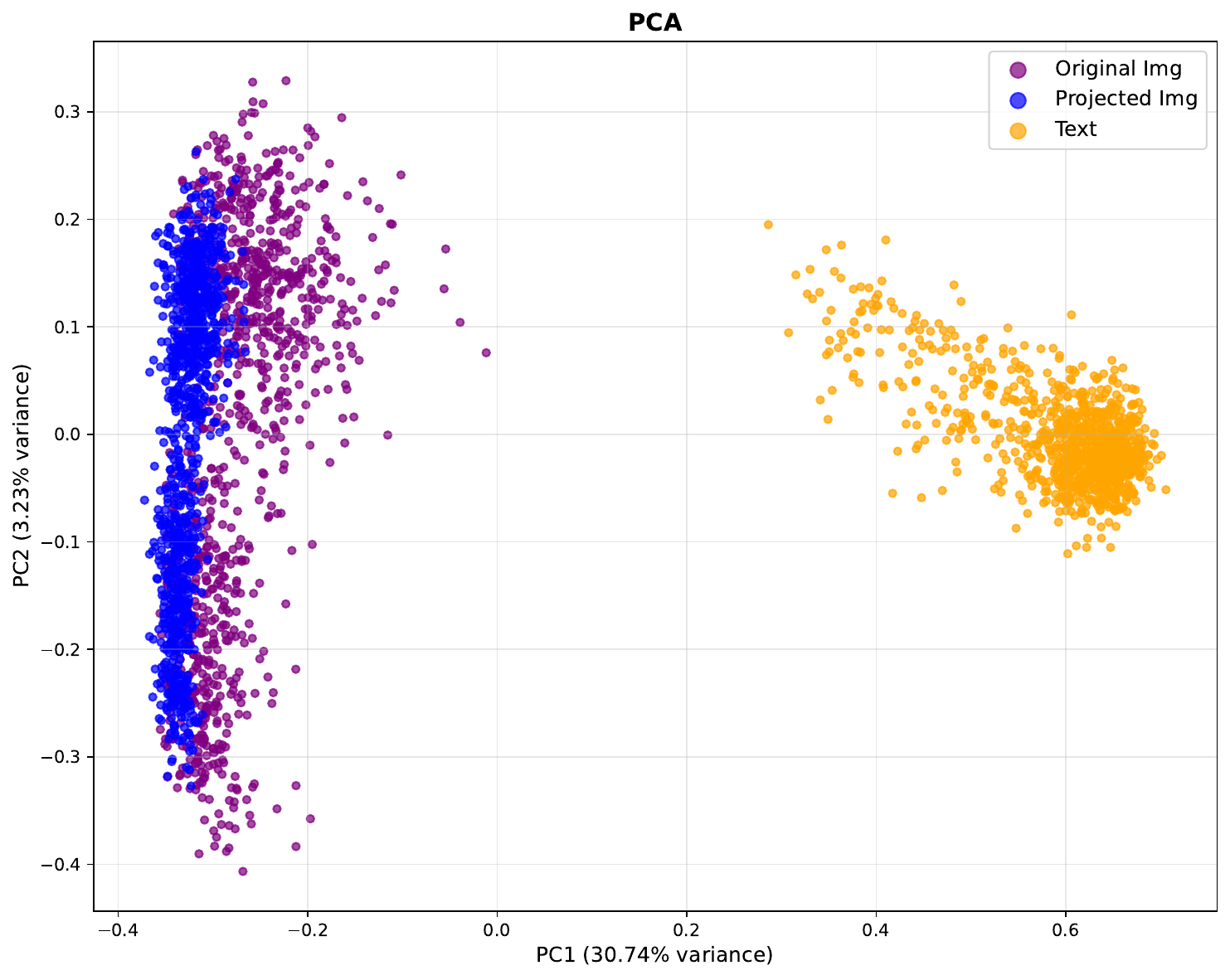}
    \end{subfigure}
    \hfill
    \begin{subfigure}{.49 \linewidth}
        \centering
        \includegraphics[width=\textwidth]{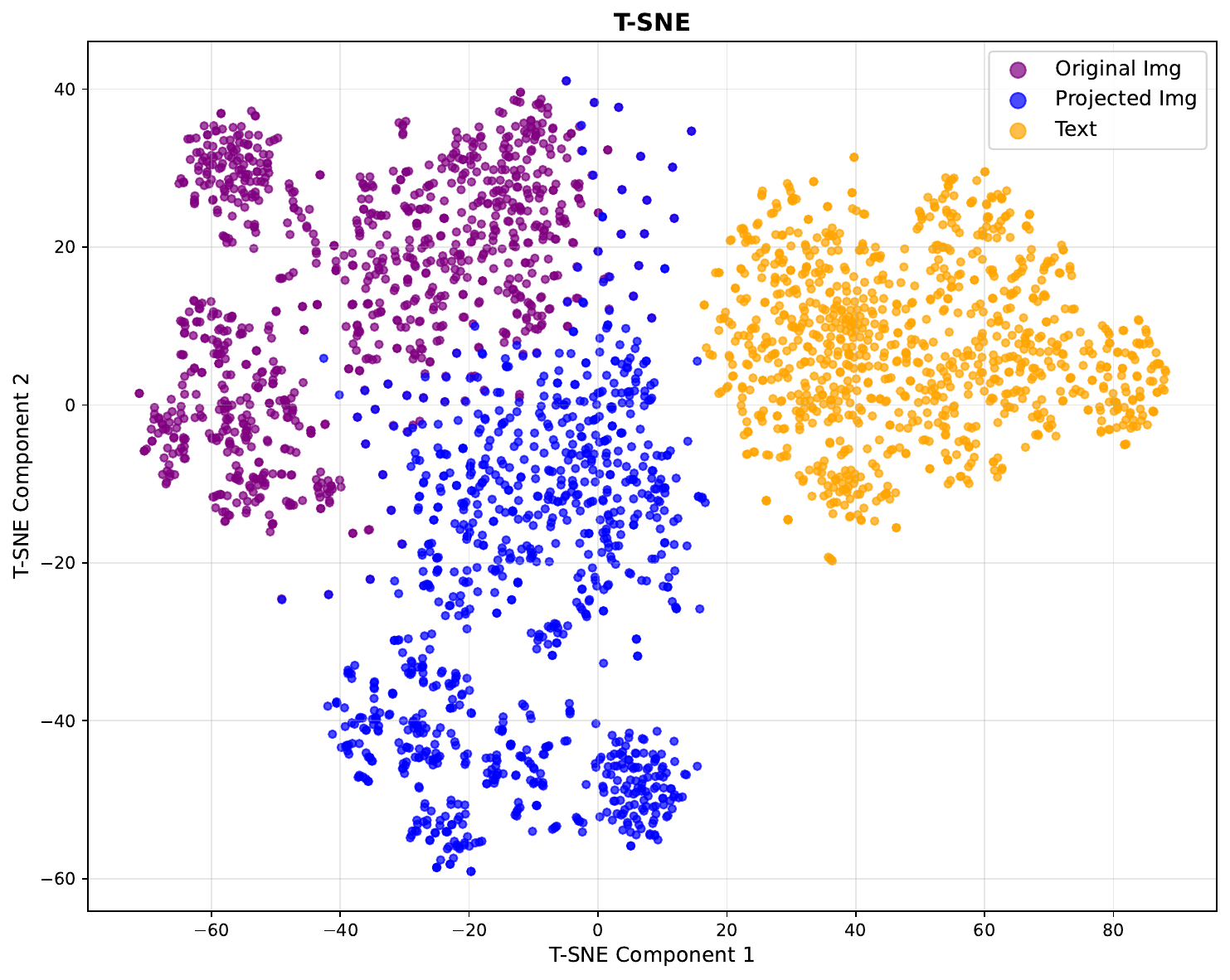}
    \end{subfigure}
    \begin{subfigure}{.49 \linewidth}
        \centering
        \includegraphics[width=\textwidth]{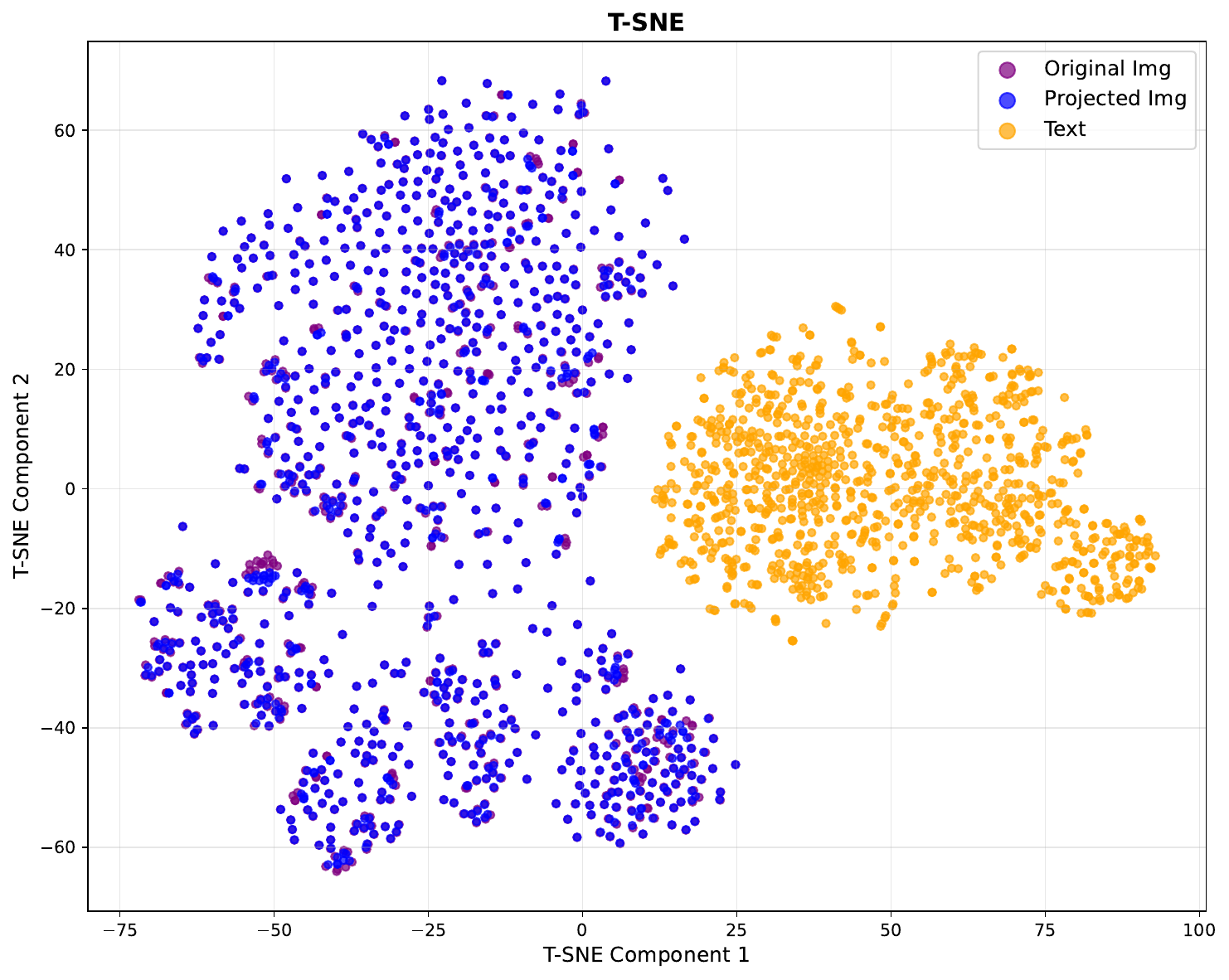}
    \end{subfigure}%
    \caption{
    %\textbf{Modality gap 1}.
    Distribution of CLIP embeddings: text (orange), original image (purple), projected image (blue). UMAP (top), PCA (middle), t-SNE (bottom).
    Left column: projection via \cref{eq:projection}. Right column: mean-adjusted projection via \cref{eq:meanadjustment}.
    \textit{Interpretation}: (i) Like many previous studies noted, text embeddings and image embeddings lie in two different cones, such that we can find them clearly separated in the figure. %often described as cones. c, or, since normalized, caps on the hypersphere.
    We argued in the main paper this ``modality gap'' does not lead to an intra-modal misalignment.
    (ii) Projecting via $PCA^\leftarrow$ and re-projecting introduces a global shift (see left PCA) in $\mathbb{R}^{512}$ that is orthogonal to the principal components $\in \mathbb{R}^{256}$.
    (iii) For comparison in $\mathbb{R}^{512}$, we can compensate for this shift by adding back the mean of the canceled components (right column).
    ViT-B/16, ImageNet val.
    }
    \label{fig:distributions_proj}
     %\caption{Raw CLIP and image embeddings projected (purple) on the 256 principal components with the highest text embedding variance.}
\end{figure*}

\section{On the projection for more ``classness''}
\label{sec:appdx_proj}
In Sec.\ 3.3 of the main paper, we presented a way to reduce the semantics of an image embedding to its dominant concept by projecting onto axes spanned by class names.

Here we show that despite usage of text embeddings, this method, $PCA^{\leftarrow}$, still can be considered image-image comparison.

\textbf{Interpretation} -
A neat way to illustrate this symmetry is by interpreting the projection of an image embedding $x_i \in \mathbb{R}^d$ as a sequence of three operations: a rotational change of basis ($Q^\top$) into the coordinate system defined by the principal components of class names, followed by a scaling ($\Lambda$) that preserves or cancels components, followed by the change back to the original basis ($Q$):
%A neat way to illustrate this is breaking down the process of projecting an image embedding $x_i \in \mathbb{R}^d$ into a change of basis ($Q^\top$), followed by a scaling along the new axes ($\Lambda$), followed by a change back to the original basis ($Q$):
\begin{equation}
    \label{eq:projection}
   x_i^{\leftarrow} = Q  \Lambda  Q^\top x_i.
\end{equation}
%for some image embedding $x_i \in \mathbb{R}^d$.
The resulting $x_i^{\leftarrow} \in \mathbb{R}^d$ is the projected image embedding.

The columns of $Q \in \mathbb{R}^{d \times d}$ contain the sorted eigenvectors (components) obtained by Eigendecomposition (PCA) of the covariance matrix of ImageNet class name text embeddings.
The orthogonality of $Q$ brings the isometric property that ensures $Q$ and $Q^\top$ themselves preserve angles and distances, and hence also similarities.

The diagonal $\Lambda \in \mathbb{R}^{d \times d}$ determines the scaling of each basis vector.
%Both $Q$ and $\Lambda$ are $d \times d$, $Q$ is orthogonal and contains the eigenvectors and $\Lambda$ is diagonal and contains the eigenvalues.
%In \cref{eq:projection}
If $\Lambda = I$, then the projection has no effect ($ x_i^{\leftarrow} = x_i$). 
We can instead set $\Lambda=\text{diag}(1...,1,0,...,0)$ to eliminate the components that do not explain much variance of class names.
After (optional) re-projection to the original space via $Q$, the resulting $x_i^{\leftarrow}$ can be interpreted as the original $x_i$ being preserved in selected directions, while cut off in the other.

\begin{figure*}
    \centering
    \includegraphics[width=\linewidth]{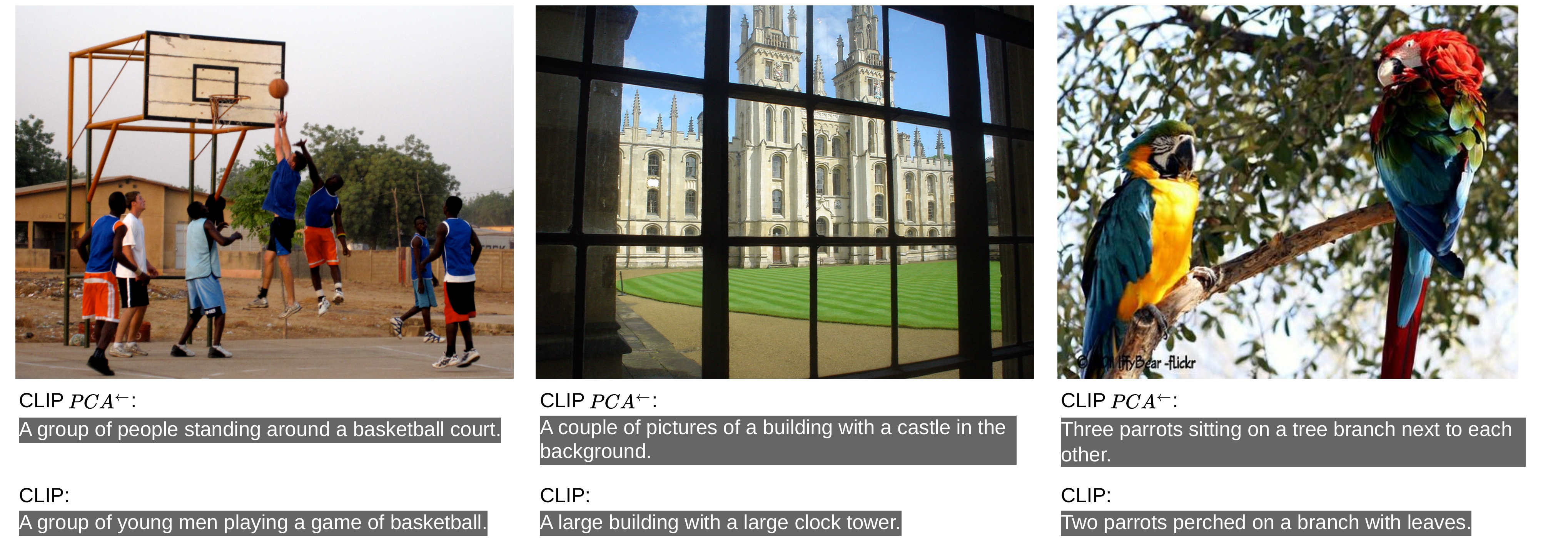}
    \caption{Captions generated with CLIP Prefix Captioning (Mokady \etal 2021), CLIP ViT-B/32, GPT-2. Swapping out the original CLIP image embedding $x_i$ for our projected $x_i^\leftarrow$, we can observe the captions still cover the main objects.
    %However, some relational details are wrong (standing around vs playing; three vs two parrots), which could be attributed to the information lost in the projection. 
    Samples are from datasets used for retrieval and few-shot classification in the main paper, left to right: SUN, ROxford, ImageNet.}
    \label{fig:captioning}
\end{figure*}

\textbf{Visualizations} with UMAP, t-SNE and PCA in \cref{fig:distributions_proj} demonstrate that $x_i^{\leftarrow}$ is indeed both globally and locally close to $x_i$.
For visualization purposes, mean adjustment can be optionally performed via $\tilde{x_i}^\leftarrow = x_i^\leftarrow + \delta_{\mu}$,
\begin{equation}
    \delta_{\mu} = Q(1-\Lambda)Q^\top \mu_x,
    \label{eq:meanadjustment}
\end{equation} i.e. adding back the part of the image embedding mean $\mu_x$ that was cut off during projection in \cref{eq:projection}.
Since this is a translation, orthogonal to all $x_i^\leftarrow$, it has no effect on distances between $x_i^\leftarrow$; it only shifts them back towards the original center $\mu_{x}$ (compare  \cref{fig:distributions_proj} left and right).
%hence also retrieval or few-shot classification performance unchanged.
Besides these plots, \cref{fig:captioning} shows how CLIP-conditioned captions can still be generated with projected $x_i^\leftarrow$.

\textbf{Concluding}, it appears valid to interpret comparison of two $x_i^{\leftarrow}$ still as image-image comparison.
Decent performance from this $PCA^{\leftarrow}$ comparison then suggests there is no issue with an intra-modal misalignment.

%\paragraph{Shortcut in Implementation.}
%In implementation, one does not need to follow the above steps that were constructed to provide some intuition.
%More similar to conventional PCA, two projected embeddings can be equivalently compared in the lower-dimensional (e.g. $\mathbb{R}^{d/2}$) subspace formed by the first selected columns in $Q$, without the need to back-project into $\mathbb{R}^d$.

\begin{figure*}
    \centering
    \begin{subfigure}{.32 \linewidth}
        \centering
        \includegraphics[width=\textwidth]{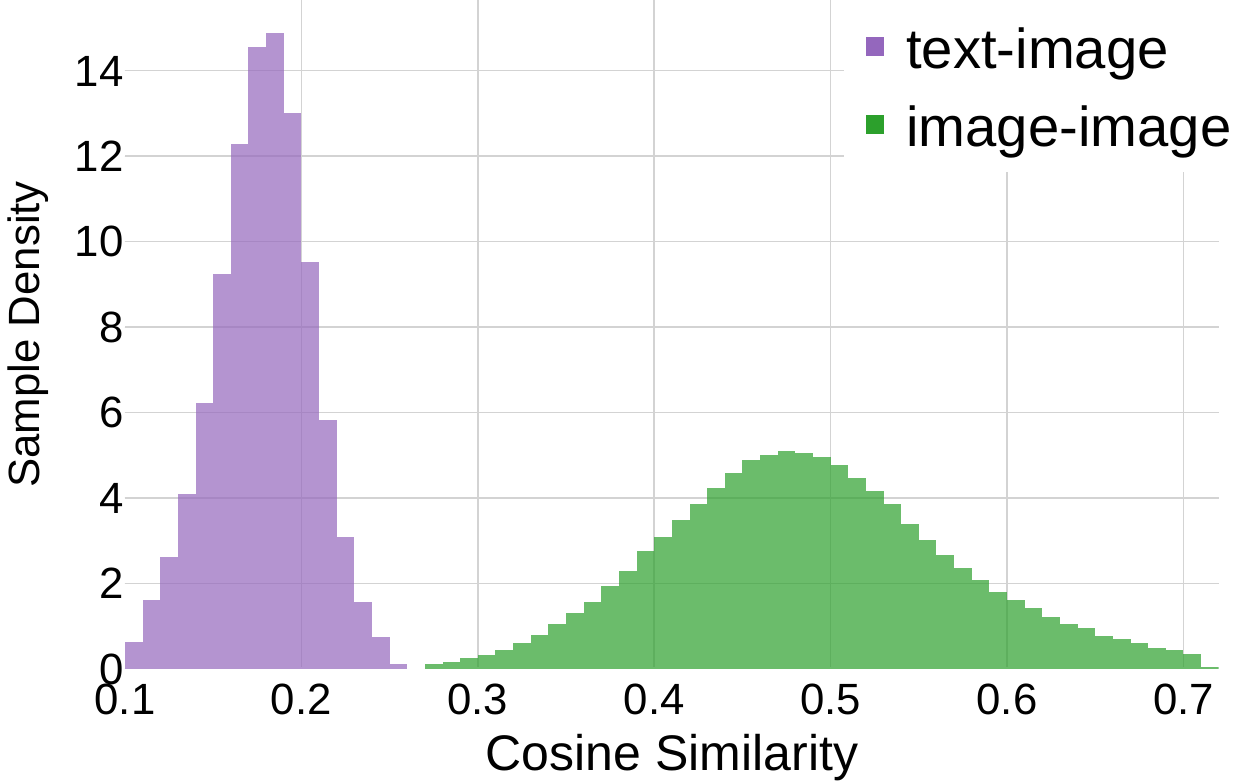}
    \end{subfigure}
    \begin{subfigure}{.32 \linewidth}
        \centering
        \includegraphics[width=\textwidth]{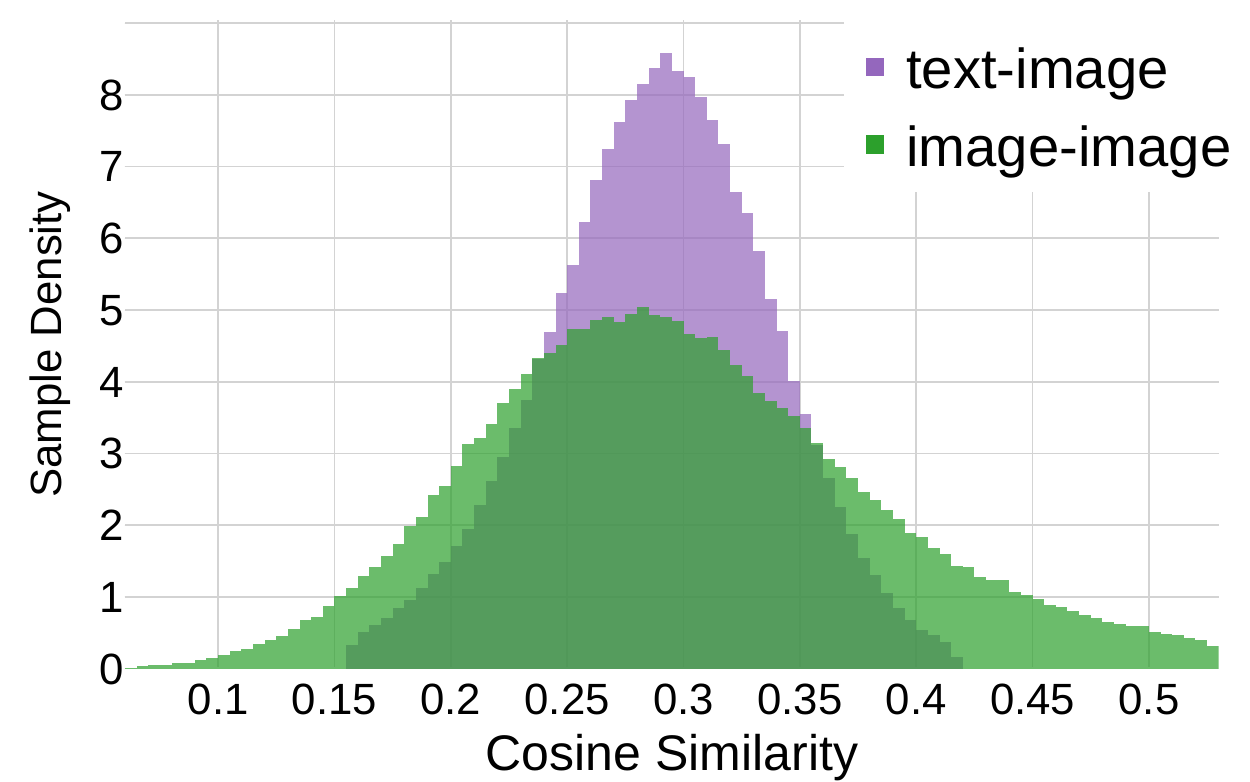}
    \end{subfigure}
    \begin{subfigure}{.32 \linewidth}
        \centering
        \includegraphics[width=\textwidth]{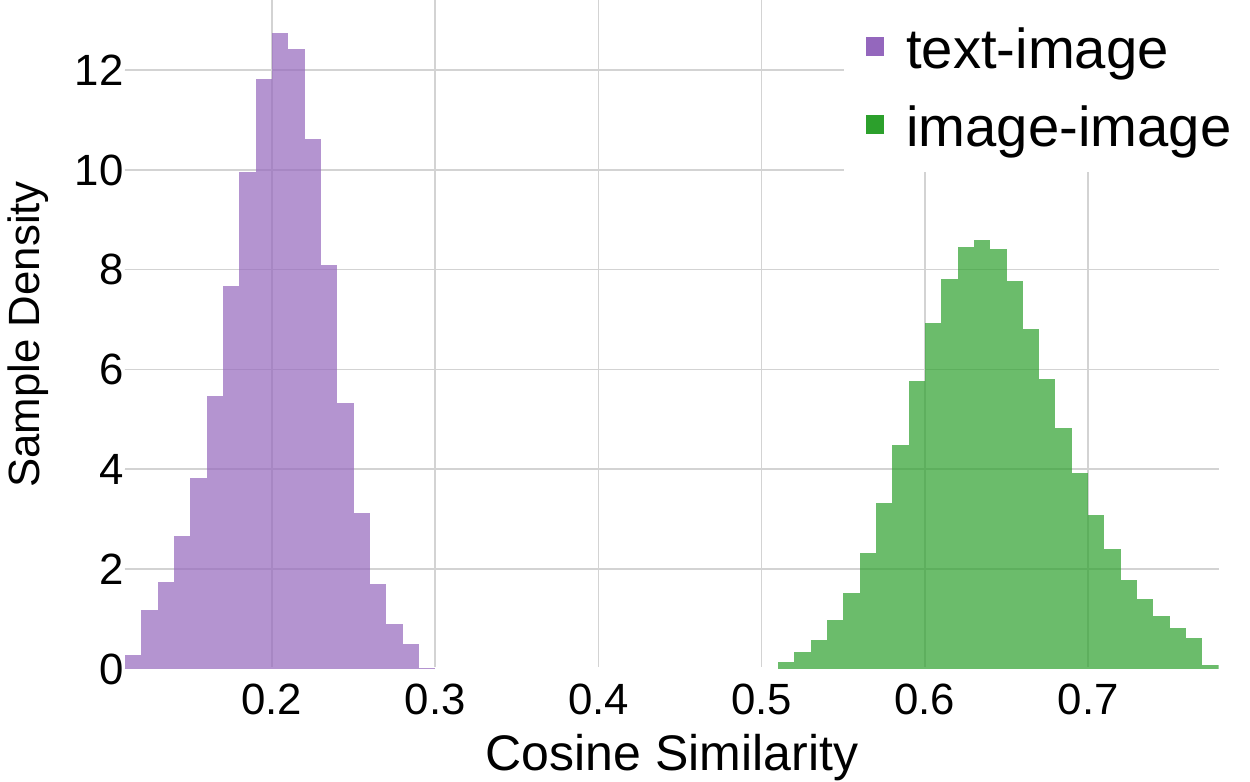}
    \end{subfigure}
    \caption{
    %\textbf{Modality gap 2}.
    Modality gap:
    %Original CLIP (left), projected (middle), mean-adjusted projected (right).
    In the original CLIP space, image-image cosine similarities are high (left).
    With the projection from \cref{eq:projection}, these similarities seem to decrease (middle).
    Applying mean adjustment as in \cref{eq:meanadjustment} removes this effect (right).
    \textit{Interpretation}: Because mean adjustment preserves Euclidean distances, the observed changes arises solely from normalization:
    the projection zeros out components, reducing vector norms such that embeddings lie no more on, but inside the unit hypersphere.
    Normalization back on the hypersphere then squeezes the projected embeddings apart, leading to a lower cosine similarity value range (middle) compared to original CLIP (left).
    Adding back the canceled mean before normalization avoids this phenomenon (right).
    %Since this is a translation orthogonal to all $x_i^\leftarrow$, it keeps image-image distances unchanged, hence also retrieval or few-shot classification performance unchanged.
    At the same time, there is no significant performance change between (middle) and (right), which once more illustrates that such cosine similarity histograms are insufficient indicators of alignment quality. ViT-B/16, ImageNet validation set.}
    \label{fig:modgaphisto}
\end{figure*}

\begin{figure*}
    \centering
    \includegraphics[width=\linewidth]{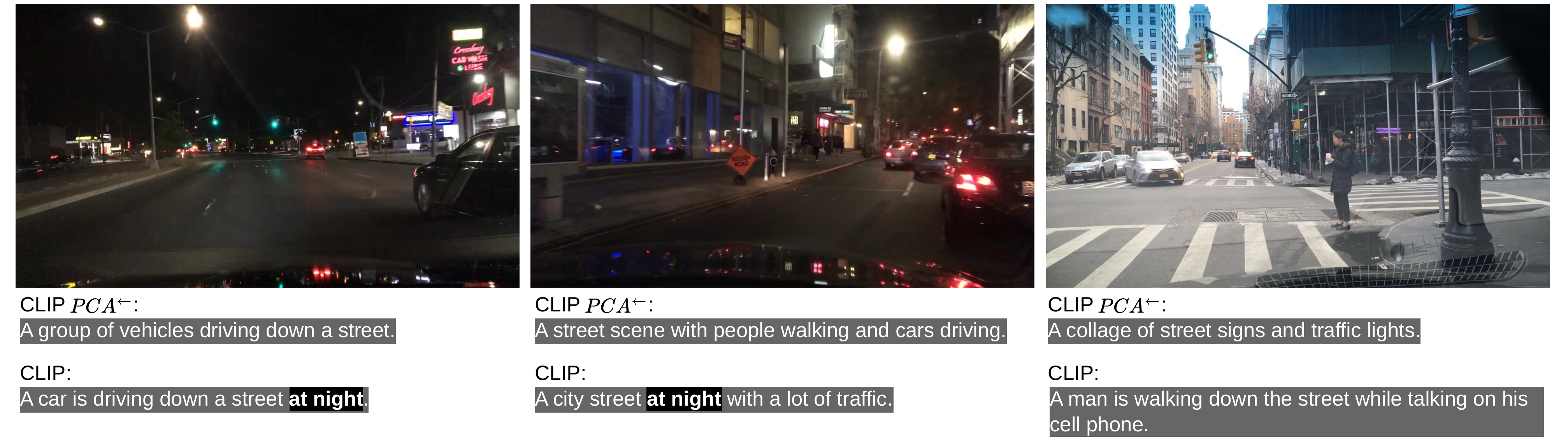}
    \caption{The same experiment as in \cref{fig:captioning}, but with the street scene dataset BDD100k used in the ablation in Sec.\ 4.3 of the main paper to validate our interpretation that the projection increases the ``classness'' of the image embedding, thereby being beneficial for classification-like tasks, but harmful for tasks such as daytime recognition because some information is lost. In line with the finding of the main paper, here we can see how the captions generated with $PCA^\leftarrow$ ignore that it is night (left, middle) and focus on the main objects only (right).}
    \label{fig:captioning_bdd100k}
\end{figure*}

\end{document}